\documentclass[3p,authoryear,11pt]{elsarticle}

\usepackage[utf8]{inputenc}
\usepackage{lmodern}      % vector Latin Modern fonts (selectable, copy-pasteable)
\usepackage[T1]{fontenc}
\usepackage{cmap}         % add ToUnicode CMap so PDF text extracts cleanly
\input{glyphtounicode}    % map glyphs (incl. ligatures) to Unicode for copy-paste
\usepackage{graphicx}
\graphicspath{{figures/}}
\usepackage{booktabs}
\usepackage{amsmath}
\usepackage{amssymb}
\usepackage{multirow}
\usepackage{array}
\usepackage{siunitx}
\usepackage{xcolor}
\usepackage{listings}
\usepackage{url}
\usepackage[hidelinks]{hyperref}

\begin{document}

\begin{frontmatter}

\title{EG-ARSA: An Expert-Grounded Open Model for Visual Road
Safety Auditing in Low-Resource Settings}

\author[ce]{Md Thamed Bin Zaman Chowdhury\corref{cor1}}
\ead{zamanthamed@gmail.com}
\cortext[cor1]{Corresponding author.}

\author[ce]{Moazzem Hossain}

\affiliation[ce]{organization={Department of Civil Engineering, Bangladesh
University of Engineering and Technology (BUET)},
            city={Dhaka},
            country={Bangladesh}}

\begin{abstract}
Road traffic injuries remain a major challenge in low- and middle-income countries,
where proactive road safety auditing is limited by incomplete crash records,
shortages of qualified auditors, and the high cost of large-scale field inspections.
These constraints make nationwide safety assessment difficult using conventional
engineering practices. To address this problem, we propose \emph{Expert-Grounded
Distillation} (EGD), a novel artificial intelligence framework that transfers
institutional road safety expertise into a compact vision-language model for
scalable visual road safety auditing. The key innovation is a quantified
expert-grounding stage in which the teacher vision-language model is calibrated
against authoritative field audits conducted by a national road safety research
institute under a World Bank--financed rural transport project. Large-scale
annotation is permitted only after the teacher reaches substantial agreement with
expert risk assessments (Cohen's $\kappa = 0.74$). The calibrated teacher then
generates structured supervision that is distilled into an 8-billion-parameter
student vision-language model using Low-Rank Adaptation and a single leakage-free
prompt. We also introduce Bangladesh Road Safety Audit (\textbf{BD-ARSA}), the
first open, expert-grounded Bangladeshi visual road safety audit dataset containing
21{,}947 image--audit records with near-national coverage, and Expert-Grounded Road
Safety Auditor (\textbf{EG-ARSA}), the first vision-language model developed
specifically for this task. Experimental results show that grounded fine-tuning
substantially improves ordinal risk assessment over the zero-shot baseline, while
blind expert evaluation demonstrates that the compact student outperforms both its
31-billion-parameter teacher and Gemini-2.5-Flash. These findings demonstrate that
EGD provides an effective and scalable engineering solution for proactive road
safety auditing in resource-constrained environments.
\end{abstract}

\begin{keyword}
Road safety audit \sep Vision-language models \sep Knowledge distillation \sep
Low-resource settings \sep Parameter-efficient fine-tuning \sep Open dataset
\end{keyword}

\end{frontmatter}

%==============================================================
\section{Introduction}
\label{sec:intro}

Road traffic crashes kill about 1.19 million people every year. Despite possessing
only about 60\% of the world's vehicles, low- and middle-income countries (LMICs)
suffer 90\% of these fatalities. Furthermore, vulnerable demographics --- specifically
pedestrians, cyclists, and motorcyclists --- comprise nearly 53\% of this global
death toll \citep{who2023}. As a lower-middle-income nation \citep{metreau2024},
Bangladesh experiences a disproportionately severe impact from this crisis.
Official figures estimate around 4{,}000 deaths and 200{,}000 injuries each year,
with the economic toll of these crashes equating to as much as 5.1\% of the
national GDP \citep{worldbank2022}. The country's victim demographic is
exceptionally vulnerable: approximately 80\% of all road fatalities involve
pedestrians, cyclists, or motorcyclists, while in Dhaka, pedestrians account for up
to 70\% of urban traffic deaths \citep{hoque2021}. Effectively mitigating these
hazards and preventing further loss of life relies entirely on accurate data.
However, in this context, the fundamental data infrastructure required to locate
and address these dangers is painfully scarce.

Conventional road-safety management is crash-based. It identifies black spots from
the historical crash record and treats them. This paradigm fails if crash data is
missing or under-reported. Road-crash data in most LMICs is scarce and unreliable:
official statistics substantially under-report road traffic injuries, with large
discrepancies between national records and global statistical estimates
\citep{mitra2023}. Bangladesh's crash-recording system has been described as
cumbersome, error-prone, and unsuitable for analysis, with a greater than 90\%
discrepancy between government recorded fatalities and WHO estimates. Moreover,
there is little usable crash data since 2016 \citep{worldbank2022}. In several
LMICs, police First Information Reports (FIRs) yield severe underreporting, as its
capture rate is as low as 13--30\% \citep{bhuiyan2022,newaz2026,rabbani2022}.
Crash-frequency and black-spot methods that work in high-income countries are
therefore unreliable here, and a different solution is needed.

Proactive infrastructure auditing can address this issue. A Road Safety Audit (RSA)
is ``a formal safety performance examination of an existing or future road or
intersection by an independent, multidisciplinary team'' \citep{fhwa2018}. RSA
scores infrastructure hazards rather than crash history, so it is robust in
comparison to gaps in the crash records. Proactive, iRAP-style star rating saves lives at scale. In a
study, an estimated 159{,}936 fatal-and-serious injuries are averted annually
across 74 countries by this method \citep{li2024}. Bangladesh has begun to
institutionalise this approach. Under the World Bank--financed Second Rural
Transport Improvement Project (RTIP-II, Additional Financing; P166295), the Local
Government Engineering Department (LGED) commissioned faculty of the Accident
Research Institute (ARI) at the Bangladesh University of Engineering and Technology
(BUET) to conduct formal field RSAs across roughly 1{,}433 km in 18 districts,
following the LGED audit methodology. The reports of this project are the sources of
this study's expert ground truth. Jurisdiction defines the road class in Bangladesh:
LGED governs rural and suburban (upazila/union) roads, the Roads and Highways
Department (RHD) governs national and regional highways, and City Corporations
govern metropolitan streets. Because our expert audits come from LGED, EG-ARSA
targets the rural/suburban road class by design, which is the segment carrying the
heaviest unaddressed rural-road crash toll.

Formal audits, however, do not scale. Globally only about 1.5 million km have been
star rated against a UN target near 12 million km by 2030 \citep{li2024}; each RSA
requires a multidisciplinary expert team on site, and the ARI--BUET programme itself
proceeded in phases over several years for a fraction of the network. Automation is
the only path to coverage. The question this paper addresses is whether modern
vision-language models (VLMs) can automate the auditor's judgement.

Existing automation falls into two families, neither of which fits the current need
(Section~\ref{sec:related} describes this). Supervised CNN/LSTM iRAP coders are
data-hungry, do not transfer across jurisdictions, and emit no interpretable audit
\citep{kacan2022,song2019}. Zero-shot VLM auditors generalise and produce language
output but are zero-shot, proprietary, street-view-only, and weak on fine-grained
detail \citep{jongwiriyanurak2024,ameen2026,garg2025,yang2024}. Crucially, none are
adapted to Bangladesh or the LGED methodology, none are open or fine-tuned, and none
release a reusable dataset.

The core premise of our approach is that domain-specific expert grounding, rather
than sheer model scale, improves performance in road safety auditing, and crucially,
this grounding can be effectively distilled into a smaller model. Without
fine-tuning, a standard smaller open-source VLM performs at near-chance levels on the
ordinal risk task (base Qwen3-VL QWK $=0.077$), and even massive frontier models or
large teacher models struggle to audit effectively without explicit grounding. This is
evidenced by a sharp performance drop in the teacher model itself in our testing.
Although the calibrated, grounded teacher agrees with expert risk judgement at
$\kappa = 0.74$, the same teacher run leakage-free is correct on only 36\% of
expert-set risk verdicts (Section~\ref{sec:results-compression}). Consequently,
standard knowledge distillation from this teacher would
simply transfer its unaided limitations to the student. To overcome this, we
introduce Expert-Grounded Distillation (EGD). In this process, the teacher model's
audit generation is strictly conditioned on authoritative field data (the ARI--BUET
on-site audits) and subsequently verified by humans. A compact, open-source student
model is then fine-tuned entirely on this expert-verified supervision. By
internalising this specialised knowledge, the student model becomes capable of
producing high-quality audits from a single image, ultimately outperforming both its
31B-parameter teacher and frontier proprietary VLMs.

We build two artifacts. The \textbf{model} is a compact, open, single-image
street-view auditor (Qwen3-VL-8B-Instruct adapted with LoRA) that outputs a
structured JSON audit (hazards with LGED category and severity, an ordinal overall
risk, and a recommendation). The \textbf{dataset}, BD-ARSA, is $\sim$22k
image--audit records across three provenance tiers (expert-gold PDF hazard crops,
expert-silver grounded audits, and street-view imagery), with location-disjoint
splits, released openly under CC~BY~4.0. To our knowledge, this dataset is the first
South-Asian, expert-grounded, LGED-schema road-safety visual-audit corpus. The
street-view tier samples 155 road corridors spanning 63 districts across all 8
administrative divisions of Bangladesh. The expert-grounded gold and silver audits
come from the 18 LGED-audited districts with field-verified ground truth. Because the ground truth is faculty-conducted field
audits under a national, World Bank--financed programme rather than crowdsourced or
analyst labels, the dataset directly answers the crash-data-quality critique above.
EG-ARSA is purpose-built for rural/suburban (LGED-class) roads. National highways
(RHD) and dense city networks (City Corporations) fall under other jurisdictions and
are addressed as future work (Section~\ref{sec:discussion}).

Our primary findings highlight the significant performance advantages of our
approach. Grounded fine-tuning increases the ordinal risk Quadratic Weighted Kappa
(QWK) by $+0.40$ compared to the zero-shot baseline, achieving approximately 0.48
alongside an exact-risk accuracy of 0.72. The statistical robustness of this
improvement is confirmed by non-overlapping bootstrap confidence intervals.
Furthermore, under identical single-image, leakage-free inference conditions, a blind
human-expert evaluation revealed that the 8B student model accurately assessed risk
81\% of the time. This substantially outperforms both the proprietary
Gemini-2.5-Flash model (58\%) and the 31B teacher model (42\%). This performance
ranking is independently validated by fully automated metrics, which yielded
corresponding scores of 0.74, 0.59, and 0.36. Finally, our evaluation framework
advances the existing iRAP-VLM literature by incorporating rigorous methodological
controls, including bootstrap confidence intervals \citep{efron1979}, a zero-shot
base ablation, ordinal QWK, a blind human panel, and a strict judge-bias control
mechanism \citep{zheng2023,panickssery2024} that evaluates the teacher leakage-free
to eliminate any unfair grounding advantage.

The compact 8B + LoRA design is single-GPU-cheap to train and runs on modest
hardware \citep{hu2022,dettmers2023}, so it can be deployed in low-resource settings
via a web application at a fraction of the cost of a formal audit. This turns a
methodological result into an LMIC public good: orders-of-magnitude cheaper audit
coverage where formal iRAP is unaffordable. The recipe (institutional-audit grounding
distilled into a compact open student) generalises to any jurisdiction with even a
small expert-audited corpus, and the same approach can extend to the national-highway
network (with RHD) and to city streets (with City Corporations) once their audit data
are available, building towards an end-to-end auditor spanning all road classes in
Bangladesh.

\paragraph{Contributions} The main contributions of this paper are:
\begin{enumerate}
  \item \textbf{Quantified expert grounding as a gated step.} We treat the teacher's
  \emph{generation} prompt as an object to be empirically validated: it is calibrated
  against authoritative institutional field audits (ARI--BUET under LGED), and its
  agreement with expert risk judgement is measured ($\kappa = 0.74$) before any
  nationwide label generation. Scaling the supervision is gated on this measurement,
  making expert grounding a verified, reported property rather than an assumed one.
  \item \textbf{Expert-Grounded Distillation (EGD), an integrated pipeline.} To the
  best of our knowledge, this is the first framework to unify (i) expert-ground-truth
  prompt calibration against authoritative institutional field audits, (ii) deliberate
  teacher--student prompt asymmetry, in which a scaffolded teacher prompt is distilled
  into a stripped, leakage-free student prompt, and (iii) confidence-tiered supervision
  spanning expert-gold, expert-silver, and large-scale street-view annotations within a
  single methodology. By grounding teacher supervision in professional field audits
  before large-scale distillation, EGD decisively outperforms conventional ungrounded
  distillation.
  \item \textbf{BD-ARSA and EG-ARSA: the first open artifacts of their kind.} We
  release the first open, expert-grounded Bangladeshi road-safety visual-audit dataset
  (21{,}947 records, 12-category LGED schema, location-disjoint splits) and the first
  VLM purpose-built for visual road-safety auditing in the region.
  \item \textbf{A demonstration that grounding beats scale, under rigorous controls.}
  In a real, high-stakes, low-resource setting, a compact 8B student surpasses both its
  31B teacher (leakage-free, 42\%) and a frontier VLM (Gemini-2.5-Flash, 58\%) at 81\%
  expert-graded risk correctness. It was evaluated with bootstrap confidence intervals,
  a zero-shot base ablation, ordinal QWK, a blind human panel, and an explicit
  judge-bias control.
\end{enumerate}

The remainder of this paper is organised as follows. Section~\ref{sec:related}
reviews related work on automated road safety assessment and on the adaptation and
evaluation of vision-language models. Section~\ref{sec:data} describes the curation
of the BD-ARSA dataset. Section~\ref{sec:method} presents the Expert-Grounded
Distillation methodology. Sections~\ref{sec:results} and \ref{sec:discussion} report
results and discuss findings, limitations, and future directions.
Section~\ref{sec:conclusion} concludes.

%==============================================================
\section{Related work}
\label{sec:related}

\subsection{Automated road safety assessment}

\subsubsection{Supervised and pre-VLM iRAP/RSA coding}
The initial push to automate road-safety assessments framed the challenge primarily
as supervised attribute coding. For example, \citet{kacan2020} developed a multi-task
ResNet/DenseNet architecture designed to jointly classify 52 distinct iRAP attributes
across 1{,}850 km of roads in Bosnia. \citet{kacan2022} subsequently advanced this
work by incorporating a Mapillary-pretrained backbone, attribute-specific recurrent
heads, and a recall-driven dynamic loss function, alongside releasing the extensive
iRAP-BH dataset. Prior to these developments, various CNN and CNN/LSTM pipelines were
utilised to map panoramic and street-view imagery directly to specific safety
features and star ratings \citep{song2019,sainju2019}. Additionally, segmentation- and
detection-based coding techniques were successfully demonstrated for AusRAP frameworks
by \citet{sanjeewani2021}, and for unconstrained roadways in South Asia by
\citet{rithish2021}. While these supervised models deliver robust accuracy, they
suffer from notable practical drawbacks. Specifically, they require massive amounts of
per-attribute labelled training data, fail to seamlessly transfer across different
auditing schemas or geographical jurisdictions, and yield rigid categorical codes
rather than generating clear, interpretable narrative audits. These limitations
strongly highlight the critical need for a more adaptable, language-capable approach
that minimises the reliance on extensive manual labelling.

\subsubsection{Vision-language models}
Vision-language models reframed the problem as visual question answering over road
imagery. The anchor work, V-RoAst \citep{jongwiriyanurak2024}, casts iRAP coding as
zero-shot VQA with proprietary models (Gemini-1.5-Flash, GPT-4o-mini) over the ThaiRAP
set and finds that VLMs generalise better than CNNs to unseen classes but underperform
on spatial and metric attributes. Its closest follow-up \citep{ameen2026} reproduces
this on Gemini-2.0/2.5 Flash with bootstrap confidence intervals and a reduced
image-only prompt, again zero-shot, street-view-only, and proprietary. Related efforts
apply VLMs to highway-scene understanding \citep{yang2025}, edge-deployable safety
detection \citep{tami2025}, and graded pavement scoring \citep{xu2025}, and several
Bangladesh-specific street-view safety studies use semantic segmentation rather than
generative VLMs \citep{hamim2024}. A parallel literature documents the empirical limits
of VLMs on street imagery: specialist self-supervised models substantially outperform
VLMs on fine-grained sign and object recognition \citep{garg2025}, and frontier VLMs
are weak on counting and geometry while strong on qualitative description
\citep{yang2024}. These limits bound what a single-image auditor can recover and
indicate where teacher-generated labels are most likely to be noisy. Across this
literature, the VLM auditors that exist are zero-shot, proprietary-API-based, and
street-view-only; none is fine-tuned, open, grounded in the Bangladesh/LGED
methodology, or paired with a reusable dataset, which is the gap EG-ARSA fills.

\subsection{Domain adaptation of vision-language models}

\subsubsection{Knowledge distillation and learning from teacher-generated labels}
Research demonstrates that compact student models can surpass significantly larger
architectures when trained on high-quality, teacher-generated supervision. For
instance, \citet{hsieh2023} showed that extracting teacher rationales as an auxiliary
training signal allows smaller models to achieve superior performance with reduced data
requirements, while \citet{liu2023} validated the use of machine-generated multimodal
supervision in visual instruction tuning. Furthermore, \citet{udandarao2025} posit that
active data curation functions as a distinct form of knowledge distillation that can
sometimes outperform traditional logit-matching. It is a finding that directly supports
our premise of utilising a curated dataset as the primary distillation mechanism.
Additionally, \citet{amin2025} detail how models trained on ``weak'' (synthetic or
teacher-generated) data can avoid model collapse and even exceed the capabilities of
the generator model, provided the training data undergoes careful curation and human
verification. However, prior distillation frameworks have typically grounded their
supervision in generic teacher outputs. The critical distinction introduced by EG-ARSA
is the deliberate anchoring of this supervision in formal, institutional expert audits.

We position EG-ARSA precisely against the mechanisms it builds on. The teacher--student
prompt asymmetry at the heart of EGD is an instance of \emph{context distillation}
\citep{askell2021,snell2022}, in which knowledge supplied to a model through an elaborate
context is internalised into its weights and later reproduced without that context;
EG-ARSA instantiates this idea for structured visual auditing, the teacher's context being
an expert-calibrated LGED audit prompt and the student deploying from a single
leakage-free prompt. Our result that an 8B student surpasses its 31B teacher is a
real-domain demonstration of \emph{weak-to-strong generalization} \citep{burns2023}, a
strong model elicited from supervision generated under weaker, unaided conditions. And
whereas recent domain-grounded synthetic-data pipelines assume or tune the generation
prompt ad hoc \citep{hsieh2023,udandarao2025,shi2025}, EGD makes expert grounding a
measured, gated step: the teacher prompt is validated against expert field audits
($\kappa = 0.74$) before it is used at scale. To the best of our knowledge, no prior work
combines (a) expert-ground-truth prompt calibration, (b) context-distillation-style prompt
asymmetry, and (c) confidence-tiered expert supervision into a single deployable auditing
pipeline. That integration, grounded in institutional field audits and released as open
artifacts, is the contribution of this work.

\subsubsection{Parameter-efficient fine-tuning and compact specialists}
Parameter-efficient fine-tuning strategies have significantly reduced the
computational barriers to model adaptation. Specifically, LoRA \citep{hu2022}
factorises weight updates into low-rank matrices, enabling the training of less than
1\% of the network's parameters without sacrificing the performance quality associated
with full fine-tuning. Expanding upon this efficiency, QLoRA \citep{dettmers2023}
incorporates 4-bit quantization, making it possible to execute this adaptation entirely
on a single GPU. Comprehensive surveys position these methodologies as cornerstones of
the broader Parameter-Efficient Fine-Tuning (PEFT) ecosystem \citep{han2024,lialin2023}.
Simultaneously, robust evidence across various domains indicates that lightly tuned
generalist models can effectively rival dedicated, resource-heavy specialists
\citep{bai2024,zhong2025,shinde2025}. Taken together, this literature firmly establishes
the technical viability of single-GPU fine-tuning within resource-constrained LMIC
environments.

\subsection{Evaluating generative vision-language models}
\label{sec:rw-eval}
The integrity of a distillation result rests on its evaluation design. LLM judges are
scalable but biased. They exhibit position, verbosity, and self-enhancement effects
\citep{zheng2023,gu2024} and they systematically prefer text from their own model
family \citep{panickssery2024}. This self-preference is precisely the circularity hazard
in our setting. Since the teacher generated the street-view labels, a same-family
automated judge would inflate the teacher's apparent quality. We therefore anchor
headline claims on a blind human-expert evaluation and run the teacher leakage-free to
remove its grounding advantage. For the ordinal Low/Medium/High risk head we adopt
quadratic weighted kappa (QWK), which penalises errors by ordinal distance
\citep{torre2018}, and we report bootstrap confidence intervals throughout
\citep{efron1979}.

EG-ARSA occupies the unfilled intersection of these strands: it is, to our knowledge,
the first fine-tuned, open VLM for visual road safety auditing, distilled from
institutional field audits (ARI--BUET/LGED) for the rural/suburban road class in an
LMIC, and evaluated with the rigour --- bootstrap intervals, a base-model ablation, blind
human review, and an explicit judge-bias control --- that the current iRAP-VLM literature
lacks. The next section describes the dataset that makes this possible.

%==============================================================
\section{Data curation: the BD-ARSA dataset}
\label{sec:data}

\subsection{Source: an institutional audit programme}
The ground truth for BD-ARSA comes from on-site Road Safety Audits conducted by
ARI--BUET faculty and commissioned by LGED under the World Bank--financed RTIP-II
Additional Financing (project P166295). Following the FHWA definition of an RSA as ``a
formal safety performance examination of an existing or future road or intersection by
an independent, multidisciplinary team'' \citep{fhwa2018}, the programme performed
formal audits across roughly 1{,}433 km of rural and suburban LGED-class roads in 18
districts, with ARI responsible for about 1{,}200 km. This institutional provenance is
the dataset's credibility anchor: the labels are field audits by a faculty/expert team
under a national programme, not crowdsourced or analyst annotations.

\subsection{The 12-category hazard taxonomy}
\label{sec:data-taxonomy}
The LGED methodology organises road hazards into the 12 categories listed in
Table~\ref{tab:categories}. To construct this taxonomy, we first consolidated 689 unique
auditor finding descriptions (208 short labels) into the canonical categories using two
independent zero-shot classifiers, Gemini-2.5-Flash and BART-large-MNLI
\citep{lewis2020}. Disagreements between the two models were treated as indicators of
ambiguity and routed for manual adjudication rather than being used to estimate taxonomy
reliability. The resulting category mapping was then validated on a human-reviewed sample
of 100 findings, achieving Cohen's $\kappa = 0.745$ (observed agreement $=0.77$),
indicating substantial agreement. This label-mapping validation is distinct from the
teacher--expert agreement reported in Sections~\ref{sec:data-tiers} and
\ref{sec:results}, which evaluates the alignment between VLM-generated risk assessments
and expert judgments. Appendix~\ref{app:nlp} documents this pipeline and the
inter-classifier statistics in full.

All 12 categories were assessed visually in the field by the expert auditors. EG-ARSA
learns these categories from the auditors' visual judgements, and we report all 12
throughout (Section~\ref{sec:results}).

\begin{table}[t]
\centering
\caption{The 12 LGED visual road-safety hazard categories, with one-line definitions.}
\label{tab:categories}
\small
\begin{tabular}{@{}p{3.2cm}p{10.0cm}@{}}
\toprule
\textbf{Category} & \textbf{Definition} \\
\midrule
\texttt{vision\_obstruction} & Sight distance blocked by trees, walls, vegetation,
shops, or structures at curves and junctions. \\
\texttt{pedestrian\_facilities} & Footpath, crossing, or walkway absent or broken
where pedestrian movement is high. \\
\texttt{shoulder\_condition} & Paved or soft shoulder absent, narrow, or ineffective;
no usable verge. \\
\texttt{roadside\_severity} & Poles, trees, or solid structures within the clear zone
($\sim$3~ft of the pavement edge). \\
\texttt{road\_markings} & Centreline, edge line, lane marking, or crossing absent,
faded, or poorly applied. \\
\texttt{intersection} & Junction geometry problems: staggered or skewed legs, poor
flaring, uncontrolled crossings. \\
\texttt{embankment\_safety} & High embankment or roadside drop-off without guardrail or
safety railing. \\
\texttt{speed\_management} & No speed reduction; missing humps, rumble strips, calming,
or speed signs. \\
\texttt{traffic\_signs} & Warning, regulatory, or guide signs missing, inadequate,
faded, or non-standard. \\
\texttt{skid\_resistance} & Polished or slippery pavement, poor surface texture, low
friction. \\
\texttt{drainage} & Poor surface drainage, standing water, waterlogging, erosion, or
inadequate drains. \\
\texttt{bus\_stoppage} & Bus stop, bay, or shelter absent or substandard; informal
carriageway boarding. \\
\bottomrule
\end{tabular}
\end{table}

\subsection{Composition and the three provenance tiers}
\label{sec:data-tiers}
BD-ARSA is organised into three provenance tiers whose names encode the certainty of
the image-to-hazard correspondence in the source reports. In the first source report,
each hazard photo is individually captioned by the auditors, giving an unambiguous
image-to-hazard label. It is the highest-confidence \textbf{gold} tier. In the second
report, locations are documented with multiple photos and a single hazard paragraph,
with no stated photo-to-hazard correspondence; we aligned each image to its hazard
finding and manually verified the alignments, yielding a reliable but lower-certainty
\textbf{silver} tier. The \textbf{street-view} tier is the largest by far: road-scene
imagery sampled nationwide from Google Street View \citep{anguelov2010} and given a full
structured audit by the teacher VLM under the calibrated, LGED-grounded prompt described
below. Because these locations carry no
per-location expert findings, their image-to-hazard correspondence rests on the
validated teacher prompt rather than on an expert report, placing the tier lowest in
the certainty hierarchy. The resulting hierarchy is gold $>$ silver $>$ street-view;
Table~\ref{tab:tiers} summarises the composition.

\begin{table}[t]
\centering
\caption{The three provenance tiers of BD-ARSA: record counts and label provenance.}
\label{tab:tiers}
\small
\begin{tabular}{@{}lrp{8.5cm}@{}}
\toprule
\textbf{Tier} & \textbf{Records} & \textbf{Label provenance} \\
\midrule
\texttt{expert\_gold}   & 343    & Single-hazard crops with auditor captions read
directly from report~1 (72 audited locations); no teacher prompt. \\
\texttt{expert\_silver} & 707    & Road-scene photos from report~2, image-aligned by us
to the expert hazard findings (alignment manually verified) and curated into full
structured audits under human review. \\
\texttt{streetview}     & 20{,}897 & Sampled road-scene imagery; full teacher-VLM
audits with no expert findings, validated against the 12-category taxonomy and
risk-boundary post-processed. \\
\midrule
\textbf{Total} & \textbf{21{,}947} & \\
\bottomrule
\end{tabular}
\end{table}

Two of the three tiers are grounded directly in the expert audits and use no teacher
model: gold carries the auditors' own per-photo captions, and silver consists of
road-scene photos that we aligned to the expert hazard findings (alignment manually
verified) and curated into full structured audits under human review. The street-view
tier is the data-generation half of Expert-Grounded Distillation: a 31B teacher VLM
(\texttt{gemma-4-31b-it}; \citealp{gemma2026}) produced a full structured audit for each
of the 20{,}897
sampled locations, which carry no per-location expert findings. These outputs were
validated automatically against the 12-category taxonomy, passed through a rule-based
correction of the Medium-to-Low risk boundary, and human-reviewed on a 384-record
sample.

The expert grounding of the street-view tier is carried by the teacher's
\emph{generation} prompt, which we engineered and validated on the expert audits before
generating any street-view labels. This is the reason the expert data is indispensable
even though it is small relative to the street-view tier. Starting from the LGED
methodology (the full 12-category schema plus explicit flagging rules), we iteratively
added corrective instructions targeting \emph{this particular teacher's} own failure
modes on the real gold and silver audits, for instance, under-flagging faded markings,
missing shoulders that are invisible at street level, over-flagging bus stops and
intersections, and confusing hazard categories. We then measured the tuned prompt
against the expert ground truth, where the prompt-tuned, LGED-grounded teacher reaches
risk agreement $\kappa=0.74$ (the \emph{teacher--expert} $\kappa$, the teacher reference
used in Section~\ref{sec:results}, and distinct from the label-mapping agreement
$\kappa=0.745$ of Section~\ref{sec:data-taxonomy}). Only once the prompt corrected the
teacher's weaknesses to this degree did we apply it at scale. The expert-grounded prompt
is therefore the conduit through which the institutional field audits shape the
street-view labels: without the gold and silver expert audits we could not have built
it, and the tier's reliability rests on a generation prompt whose agreement with expert
judgement was quantified on that expert data. The prompt is grounded in the LGED
methodology and the weakness-correcting rules only, with no per-location findings.

Each record is stored as a compact JSON object with fields \texttt{record\_id},
\texttt{source}, \texttt{image}, \texttt{hazards} (a list of \{\texttt{name},
\texttt{category}, \texttt{observation}, \texttt{severity}\}),
\texttt{overall\_risk\_level}, \texttt{recommendation}, \texttt{tasks\_available}, and
\texttt{split}. The \texttt{tasks\_available} field records which supervision targets a
record carries and drives the per-record loss masking described in
Section~\ref{sec:method}. For street-view records the object additionally stores the
capture latitude, longitude, heading, and Google Street View panorama identifier. The
street-view imagery is \copyright~Google and is not redistributed: accessed through the
Google Maps Platform under its Terms of Service, the street-view tier is released as
these annotations, coordinates, and panorama identifiers only, from which the images are
reconstructed locally via the official Street View Static API; the expert-gold and
expert-silver images derive from the ARI--BUET/LGED reports and are shared with LGED's
permission.

\subsection{Splits and statistics}
\label{sec:data-splits}
Splits are location-disjoint. So, no location ever crosses splits. This gives train
16{,}082 / validation 2{,}418 / test 3{,}447 records (Table~\ref{tab:splits}). The
overall-risk distribution is markedly imbalanced. Low risk is only about 1.4\% of the
training set (225 records), against 6{,}340 Medium and 9{,}517 High. This imbalance
motivates the handling described in Section~\ref{sec:method-loss}. As an
internal-consistency check, overall risk tracks the count of High-severity hazards: the
rule ``0 High-severity hazards $\rightarrow$ Low, exactly 1 $\rightarrow$ Medium,
$\geq 2 \rightarrow$ High'' matches the assigned risk in 72.2\% of records.

\begin{table}[t]
\centering
\caption{Location-disjoint splits by source tier and the overall-risk distribution per
split.}
\label{tab:splits}
\small
\begin{tabular}{@{}lrrrr@{}}
\toprule
& \textbf{Train} & \textbf{Val} & \textbf{Test} & \textbf{Total} \\
\midrule
\multicolumn{5}{@{}l}{\emph{By source tier}} \\
\texttt{expert\_gold}   & 107      & 78       & 158      & 343 \\
\texttt{expert\_silver} & 507      & 96       & 104      & 707 \\
\texttt{streetview}     & 15{,}468 & 2{,}244  & 3{,}185  & 20{,}897 \\
\textbf{All sources}    & \textbf{16{,}082} & \textbf{2{,}418} & \textbf{3{,}447} & \textbf{21{,}947} \\
\midrule
\multicolumn{5}{@{}l}{\emph{By overall-risk level}} \\
Low    & 225 ($\sim$1.4\%) & 77       & 102      & 404 \\
Medium & 6{,}340           & 1{,}100  & 1{,}381  & 8{,}821 \\
High   & 9{,}517           & 1{,}241  & 1{,}964  & 12{,}722 \\
\bottomrule
\end{tabular}
\end{table}

\begin{figure}[t]
\centering
\includegraphics[width=\linewidth]{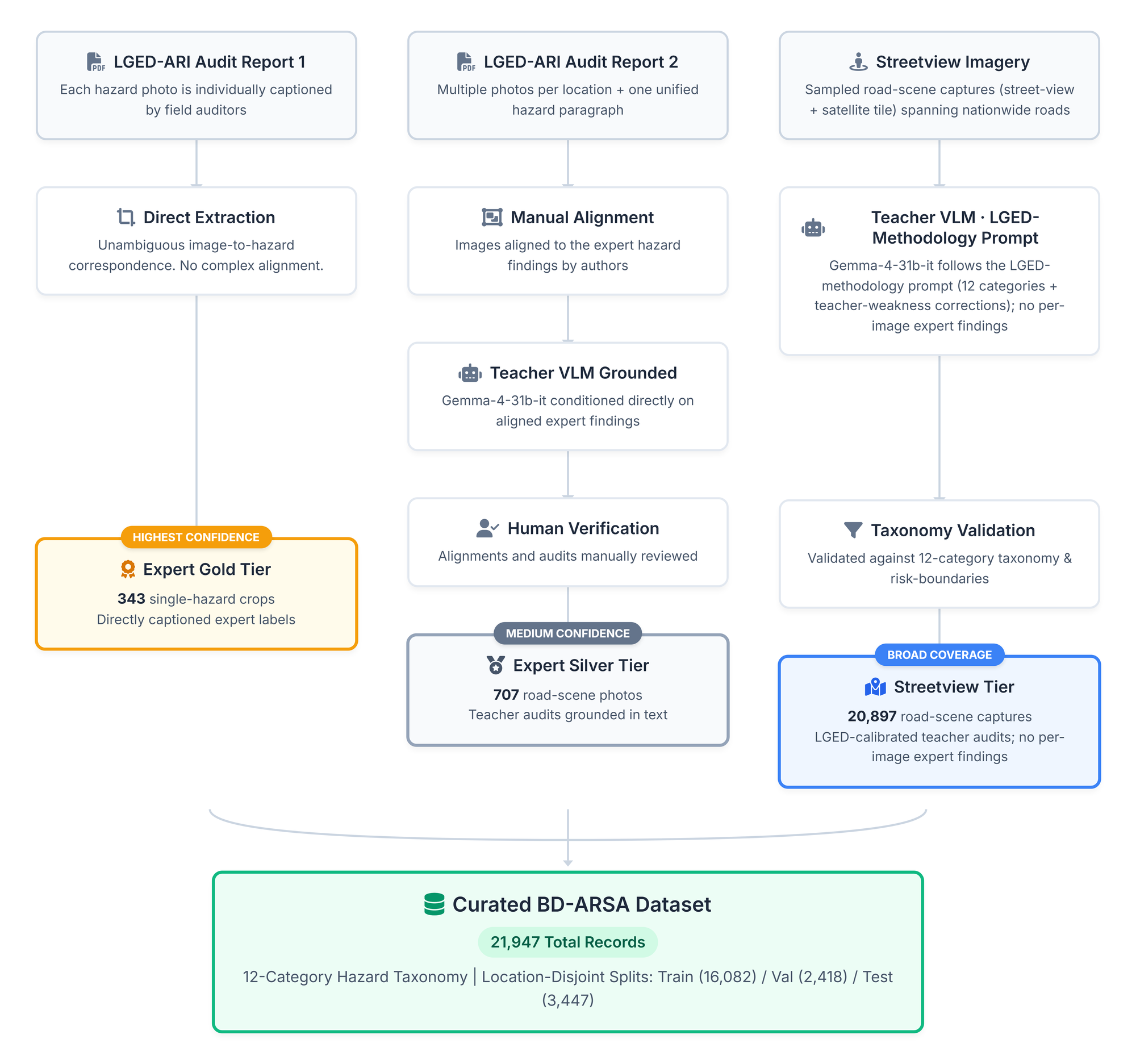}
\caption{Three-tier data provenance of BD-ARSA. Expert field audits from the
ARI--BUET/LGED programme yield the gold (individually captioned hazard crops) and silver
(road scenes we aligned to the expert findings and curated under human review) tiers;
the street-view tier is teacher-audited road-scene imagery whose generation prompt was
calibrated on the expert audits, with no per-image expert findings.}
\label{fig:provenance}
\end{figure}

\begin{figure}[t]
\centering
\includegraphics[width=\linewidth]{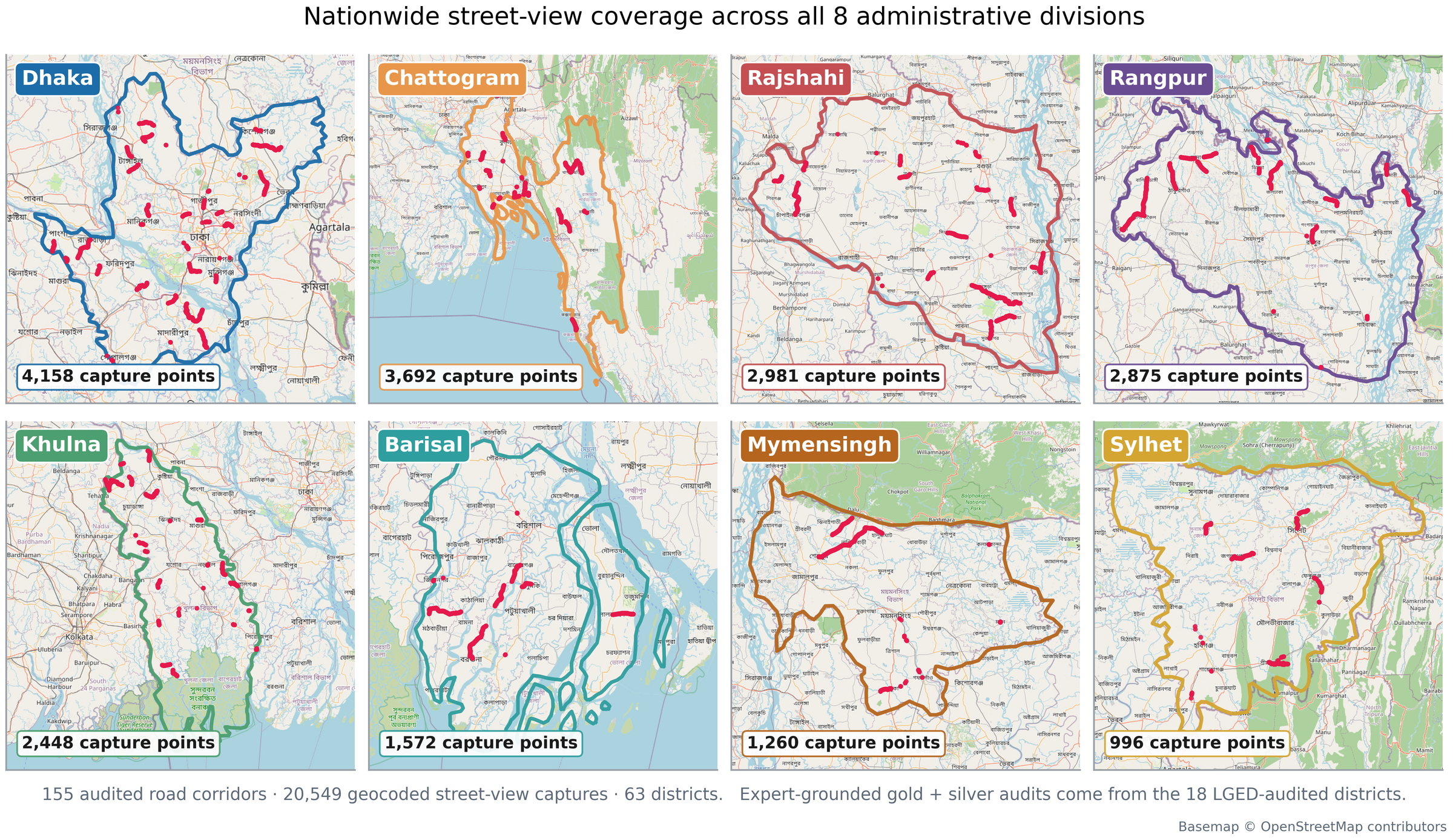}
\caption{Nationwide street-view coverage. The 155 sampled road corridors span 63
districts across all 8 administrative divisions of Bangladesh. The expert-grounded gold
and silver tiers are drawn from the 18 LGED-audited districts within this footprint.}
\label{fig:coverage}
\end{figure}

\begin{figure}[t]
\centering
\includegraphics[width=\linewidth]{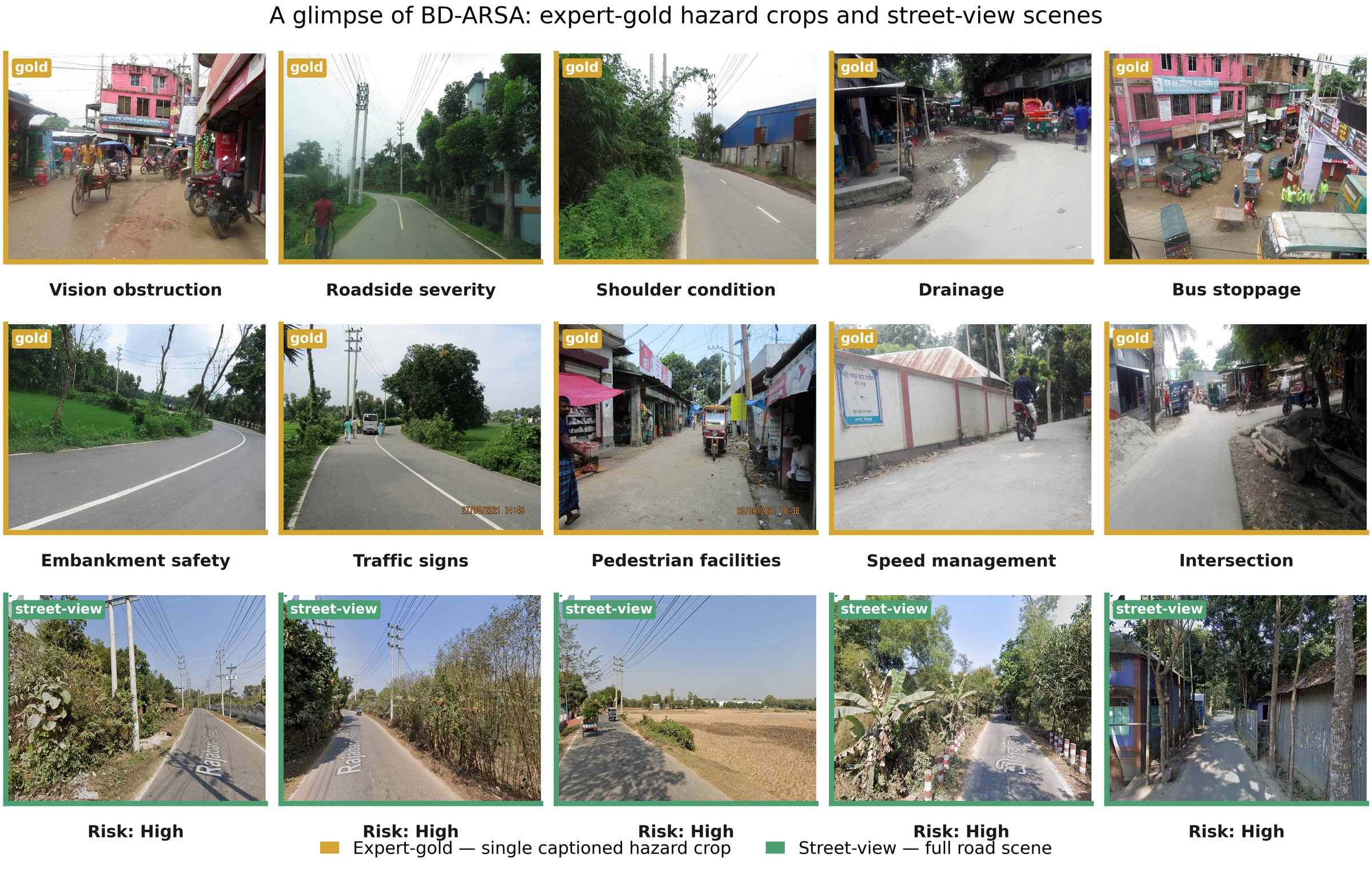}
\caption{A sample of BD-ARSA imagery, mixing gold single-hazard crops with street-view
and silver road scenes across hazard categories and risk levels.}
\label{fig:montage}
\end{figure}

Figure~\ref{fig:provenance} summarises the three-tier provenance. The street-view tier
samples 155 named road corridors (21{,}266 coordinate points) spanning 63 districts
across all 8 administrative divisions of Bangladesh, near-national coverage
(Fig.~\ref{fig:coverage}), and Fig.~\ref{fig:montage} shows a sample grid spanning
hazard categories and risk levels.

\subsection{Dataset analysis}
\label{sec:data-analysis}
\begin{figure}[t]
\centering
\includegraphics[width=\linewidth]{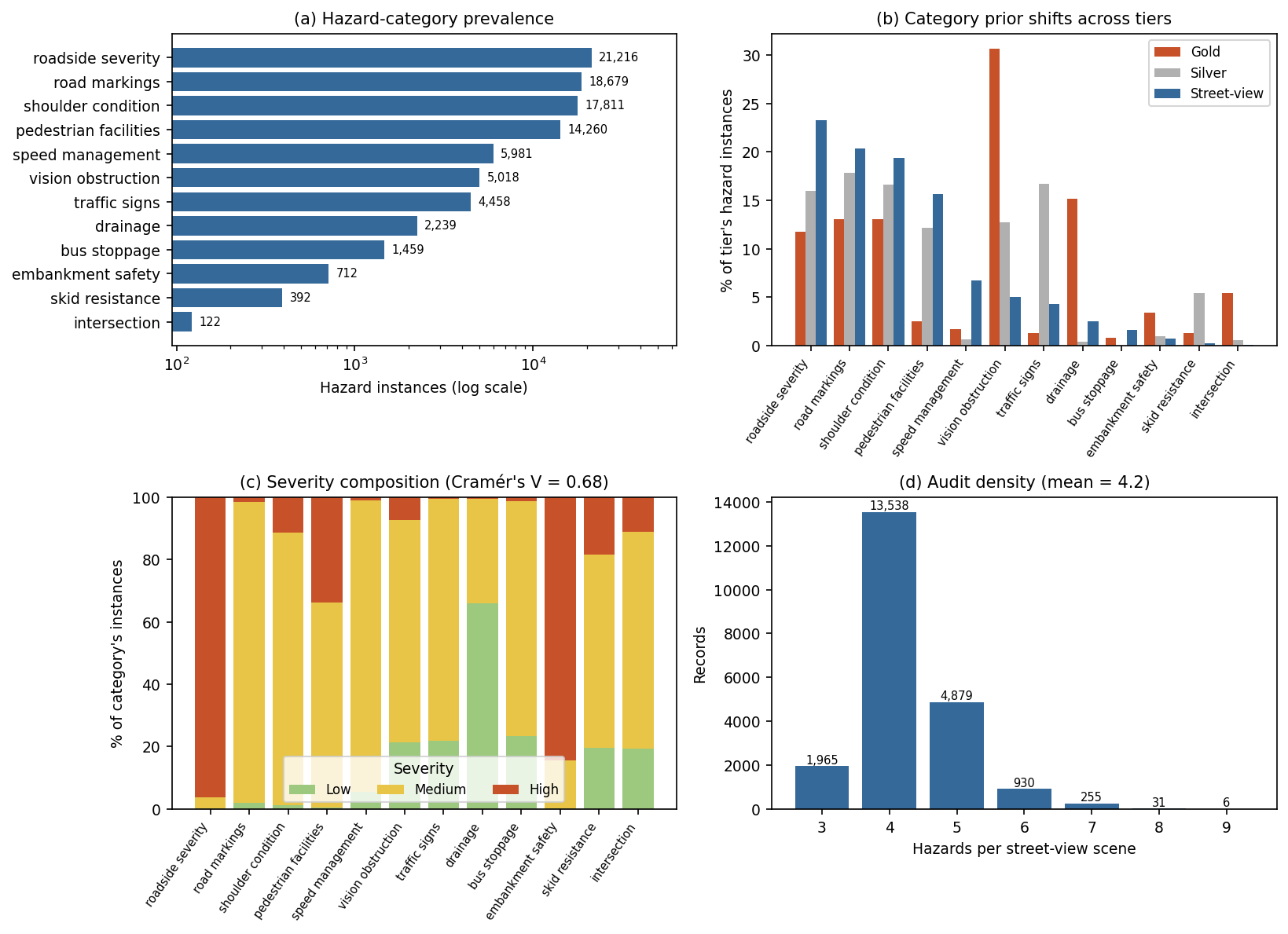}
\caption{Distributional structure of BD-ARSA. (a) Hazard-category prevalence (log scale);
(b) category composition within each provenance tier; (c) severity composition per
category (Cram\'er's $V = 0.68$); (d) number of hazards per street-view scene (mean 4.2).}
\label{fig:dist}
\end{figure}

To explicitly define the statistical structure of the BD-ARSA dataset, a quantitative
analysis was conducted (Table~\ref{tab:corpus}). The dataset's 21{,}947 records contain a
total of 92{,}347 hazard instances. With an average of 4.22 hazards per street-view image
(ranging from 3 to 9; Fig.~\ref{fig:dist}d), each record represents a complex,
multi-hazard audit rather than a single isolated label.

\begin{figure}[t]
\centering
\includegraphics[width=\linewidth]{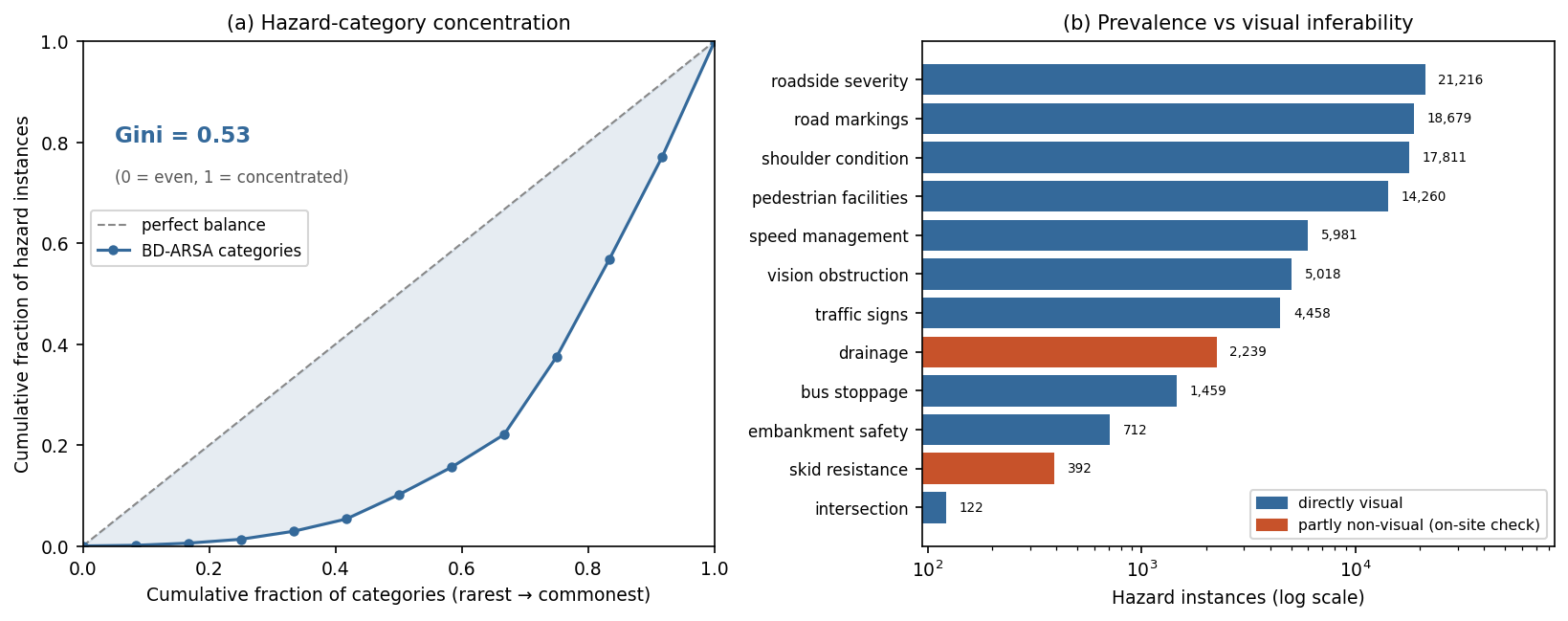}
\caption{Hazard prevalence as a base-rate prior. (a) Lorenz curve of the 12-category
instance distribution (Gini 0.53). (b) Per-category prevalence coloured by visual
inferability; the two partly-non-visual categories (\texttt{drainage},
\texttt{skid\_resistance}) --- the model's hardest set --- sit at mid-range frequency,
decoupling category difficulty from class scarcity.}
\label{fig:baserate}
\end{figure}

The prevalence of hazard categories exhibits a long-tailed distribution
(Fig.~\ref{fig:dist}a, Fig.~\ref{fig:baserate}a), with a Gini concentration coefficient
of 0.53. Categories like \texttt{roadside\_severity}, \texttt{road\_markings},
\texttt{shoulder\_condition}, and \texttt{pedestrian\_facilities} are highly dominant,
while \texttt{intersection}, \texttt{skid\_resistance}, and \texttt{embankment\_safety}
appear infrequently.

Rather than being a flaw in sampling, this skew accurately mirrors the reality of
Bangladeshi rural roads, where missing shoulders and roadside obstructions are pervasive,
while specific intersection or embankment issues are geographically localised. Because
this prevalence ordering remains stable across all three tiers of data provenance
(Fig.~\ref{fig:dist}b), it is treated as a realistic empirical prior that should be
preserved. Artificial balancing was avoided to prevent distorting the audit process and
artificially increasing false positives for rare hazards, which is a crucial
consideration for a practical screening tool. Consequently, imbalance mitigation
techniques are only utilised where a single discrete decision must be made under skew,
the ordinal risk head (Section~\ref{sec:method-loss}), and not for the
auto-regressively generated hazard lists where fixed per-class reweighting is not
well-defined (Section~\ref{sec:method-loss}).

Hazard severity is inherently tied to its specific category, demonstrating a strong
correlation (Cram\'er's $V = 0.68$; Fig.~\ref{fig:dist}c). For example,
\texttt{roadside\_severity} is deemed ``High'' risk in roughly 95\% of cases, whereas
\texttt{road\_markings} and \texttt{shoulder\_condition} issues are largely classified as
``Medium''. This indicates that the dataset's labels capture severity structures
intrinsic to the category itself, rather than applying severity as an independent,
free-floating tag.

\begin{figure}[t]
\centering
\includegraphics[width=0.7\linewidth]{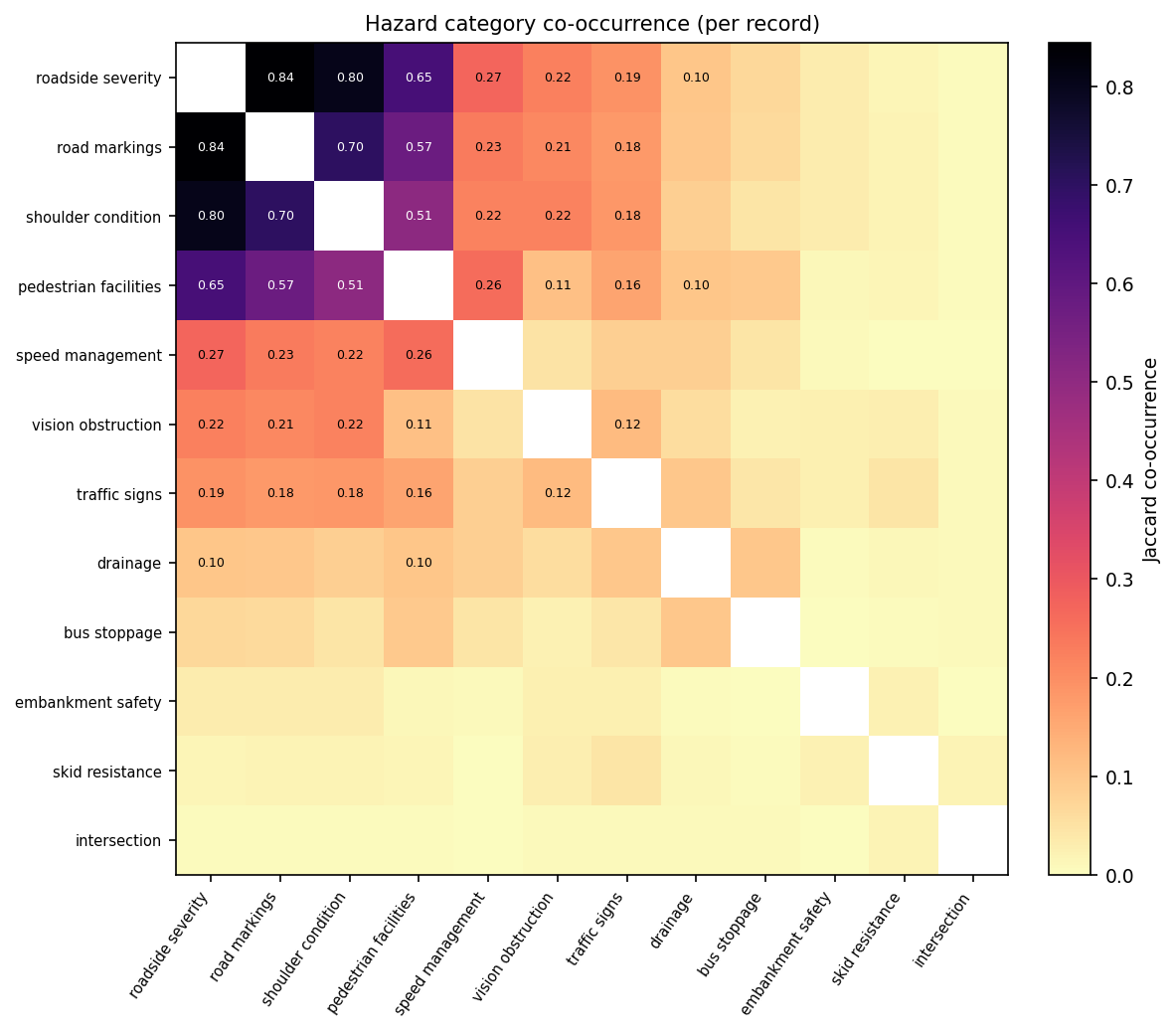}
\caption{Hazard-category co-occurrence within records (Jaccard index). The four common
categories form a tightly coupled cluster (up to 0.84); the rare categories are largely
independent.}
\label{fig:cooccur}
\end{figure}

Furthermore, hazards frequently appear in structured, co-occurring clusters
(Fig.~\ref{fig:cooccur}). The four most common categories are tightly interrelated
(reaching a Jaccard index of up to 0.84 for
\texttt{roadside\_severity}--\texttt{road\_markings}), illustrating a compound failure
scenario typical of unmarked rural roads lacking shoulders and suffering from roadside
encroachment. Conversely, the rare categories are comparatively independent. Therefore,
successful models must learn to predict these correlated hazard sets rather than treating
labels as isolated, independent variables.

Certain categories, specifically \texttt{drainage} and \texttt{skid\_resistance}, are
uniquely challenging because human auditors assess them partly from on-site, non-visual
checks (Fig.~\ref{fig:baserate}b). For a single-image auditor, these represent the
model's hardest set (Section~\ref{sec:lim}), generally only recoverable in highly obvious
scenarios.

Crucially, this difficulty stems directly from their lack of visual inferability, not a
lack of training frequency. Both categories exist at a mid-range prevalence, whereas the
absolute rarest category (\texttt{intersection}) is visually distinct and is not among
the weak set. Recognising that performance limitations on these specific hazards are tied
to the constraints of the single-image modality, rather than class scarcity, is a
distinction that matters because it confirms that simply applying category-level
rebalancing to the training data would not be expected to improve them.

Finally, despite utilising a location-disjoint method to separate the data, the held-out
test set remains a highly faithful sample of the training distribution. The distributions
for both hazard categories and overall risk levels are nearly identical between the train
and test splits (Cram\'er's $V = 0.03$ and 0.05, respectively; Table~\ref{tab:corpus}).
This consistency ensures that the model's evaluation is not confounded by distribution
shifts introduced by the disjoint splitting strategy (Section~\ref{sec:results}).

\begin{table}[t]
\centering
\caption{BD-ARSA corpus summary statistics (released dataset; metadata computed over
all 21{,}947 records).}
\label{tab:corpus}
\small
\begin{tabular}{@{}p{7.5cm}p{6.0cm}@{}}
\toprule
\textbf{Property} & \textbf{Value} \\
\midrule
Records (image--audit) & 21{,}947 (343 expert-gold / 707 expert-silver / 20{,}897
street-view) \\
Hazard instances (total) & 92{,}347 \\
Mean hazards per street-view scene & 4.22 (range 3--9) \\
Hazard categories & 12 (LGED schema) \\
Category entropy (normalised) & 0.79 (of 1.0) \\
Category Gini coefficient (concentration) & 0.53 \\
Category--severity coupling (Cram\'er's $V$) & 0.68 \\
Recommendation length (median / p99 / max tokens) & 33 / 48 / 61 \\
Road type & single carriageway (99.9\%) \\
Land use & mixed (70\%), agricultural (27\%) \\
Geographic coverage & 155 corridors, 63 districts, 8 divisions \\
Split representativeness, train vs test (Cram\'er's $V$) & risk 0.05; hazard category
0.03 \\
\bottomrule
\end{tabular}
\end{table}

%==============================================================
\section{Methodology}
\label{sec:method}

\subsection{Overview of Expert-Grounded Distillation}
\begin{figure}[t]
\centering
\includegraphics[width=\linewidth]{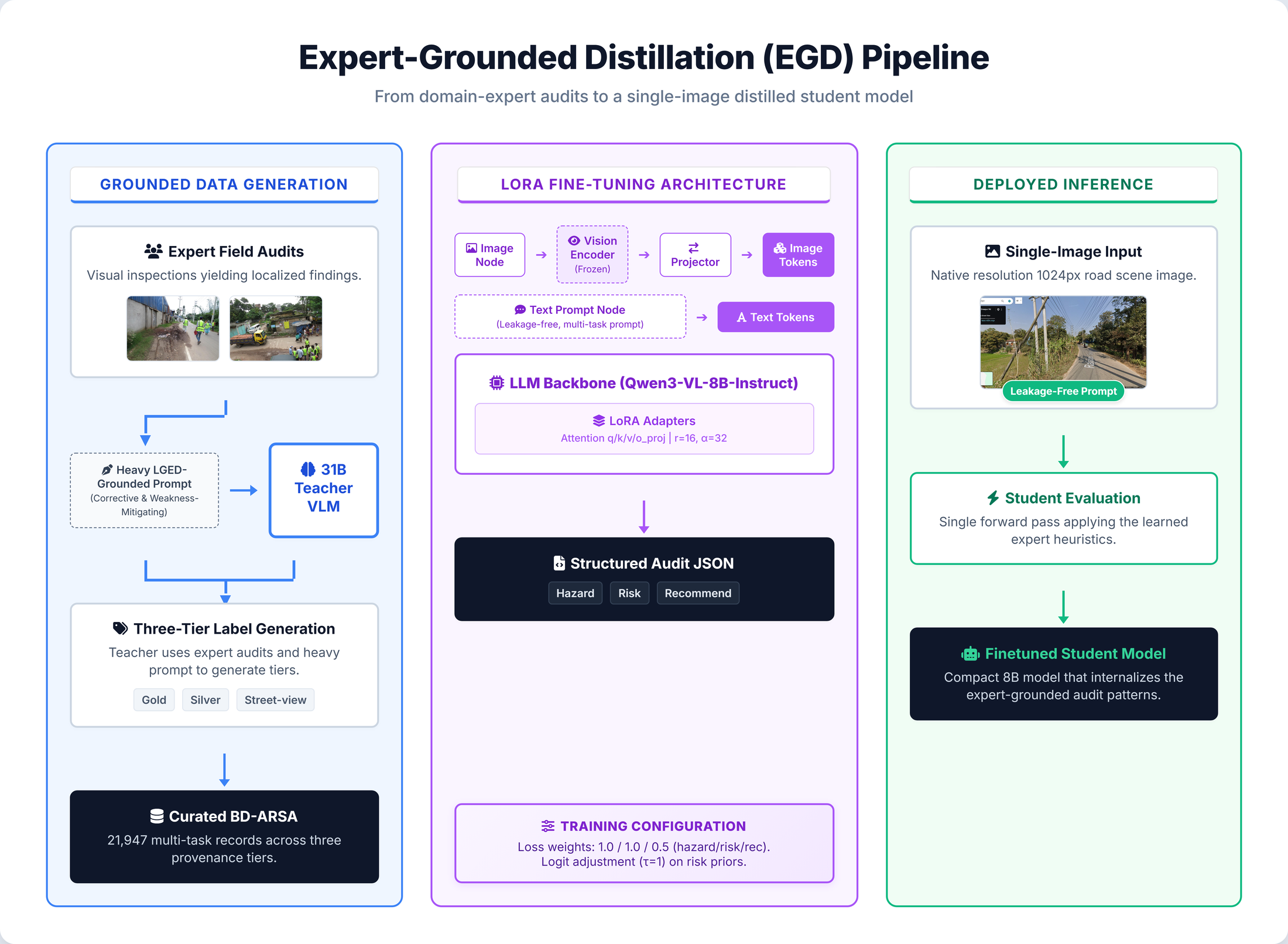}
\caption{The Expert-Grounded Distillation pipeline. \emph{Left:} grounded data generation
across the three tiers, curated into BD-ARSA (summarised here; full provenance in
Fig.~\ref{fig:provenance}). \emph{Middle:} the student architecture --- a frozen vision
encoder and projector feed image tokens, alongside text tokens, to the
Qwen3-VL-8B-Instruct backbone, which is adapted with LoRA on the attention projections
and trained with the multi-task objective and train-only logit adjustment. \emph{Right:}
single-image, leakage-free inference.}
\label{fig:pipeline}
\end{figure}

Expert-Grounded Distillation (EGD) transfers expert-grounded audit knowledge into a
compact open student. The data-generation part, which calibrates the teacher's generation
prompt on the institutional field audits and human-reviews its output, is described in
Section~\ref{sec:data}. Crucially, that calibration is a measured, gated step: we quantify
the teacher prompt's agreement with expert risk judgement ($\kappa = 0.74$) on the gold
and silver audits and apply it at nationwide scale only once it clears this bar, so the
expert grounding of the supervision is a verified property rather than an assumed one.
This section covers the training and inference side,
as well as the evaluation protocol. Figure~\ref{fig:pipeline} ties the two halves
together. Grounded supervision (gold captions, silver findings we image-aligned,
street-view teacher audits) is curated into BD-ARSA, used to LoRA-fine-tune
Qwen3-VL-8B-Instruct, and deployed on a single image at inference. The contrast with
ungrounded distillation is shown in Section~\ref{sec:results}.

\subsection{Task formulation and structured audit output}
\label{sec:method-task}
\begin{figure}[t]
\centering
\includegraphics[width=\linewidth]{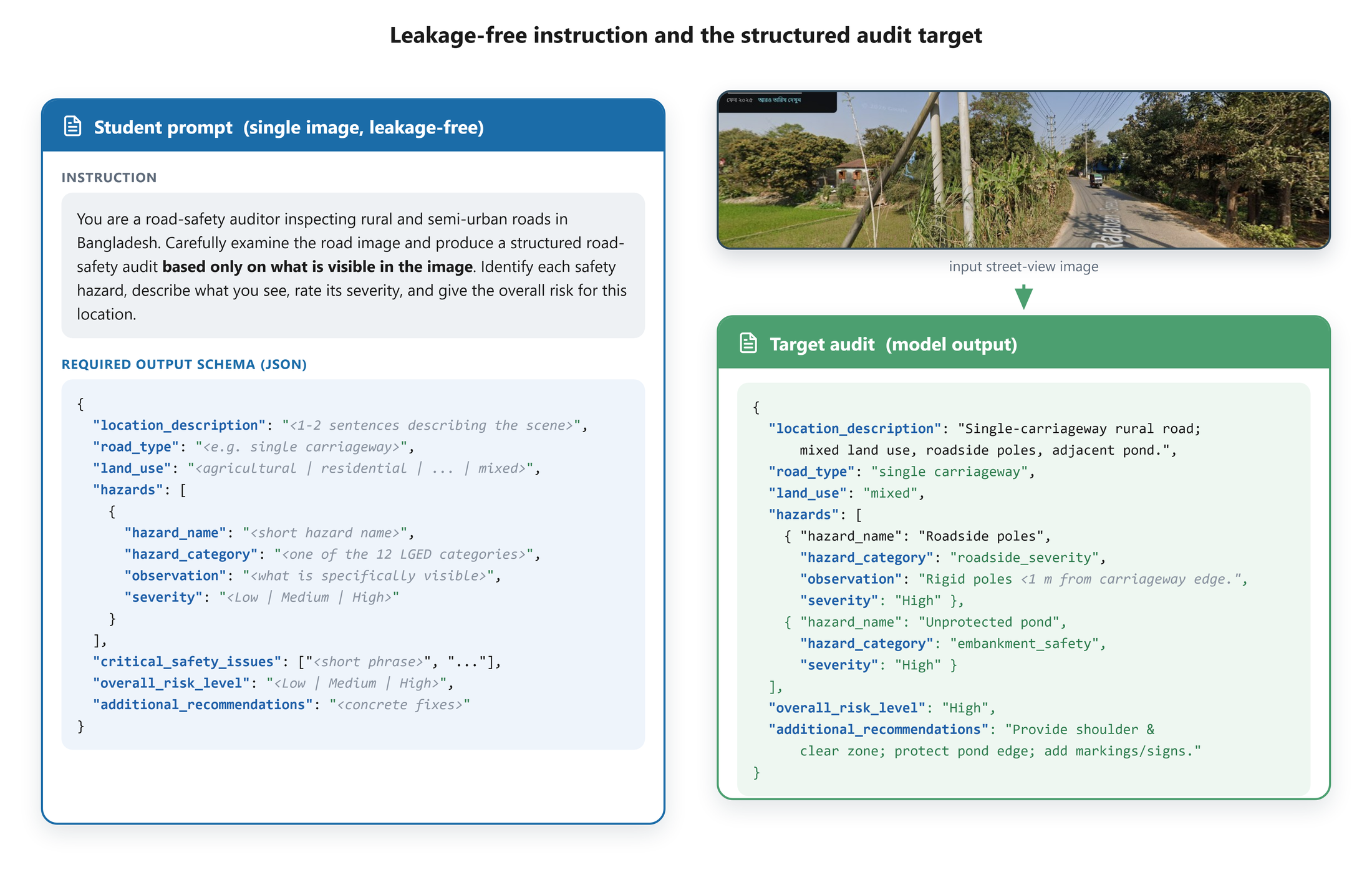}
\caption{The student prompt (left) and the structured JSON audit target (right), with a
worked street-view example. The prompt instructs the model to audit based only on what is
visible in the image; the target lists hazards with an LGED category and severity, an
ordinal overall risk, and a recommendation.}
\label{fig:prompt}
\end{figure}

Each record supervises up to three tasks: hazard generation, overall-risk classification
(ordinal Low/Medium/High), and recommendation generation. The model emits a single JSON
object in the canonical schema of Section~\ref{sec:data-tiers}.
Figure~\ref{fig:prompt} shows the student prompt and a filled target. The student prompt
is leakage-free, as it refers to no audit report or finding. So, the model is never
trained to cite a source it will not have at inference. Moreover, the student prompt also
does not contain any teacher-specific weakness mitigation instructions and is designed as
a standardised prompt.

This deliberate asymmetry in prompt design is the defining mechanism of Expert-Grounded
Distillation (EGD), and it operationalises context distillation
\citep{askell2021,snell2022} for structured visual auditing: the elaborate,
expert-calibrated context supplied to the teacher is internalised into the student's
weights and reproduced at inference without it. As detailed in
Section~\ref{sec:data-tiers}, the teacher model
generates labels using an elaborate, LGED-grounded prompt that incorporates specific
corrective instructions calibrated against the expert audits. In stark contrast, the
student model is both trained and deployed using a single, standardised, leakage-free
prompt devoid of any structural scaffolding. It relies solely on the input image without
per-category rules, grounding text, or report references. Consequently, the student
internalises the complex audit patterns directly from the training data, absorbing the
specialised knowledge that the teacher's elaborate prompts originally supplied. This
internalisation explains how the student can successfully generate audits from a simple
inference prompt, and why it outperforms its own teacher when evaluated under identical,
leakage-free conditions (Section~\ref{sec:results-compression}). That baseline comparison
strips the teacher of the critical prompt scaffolding that the student has already
encoded into its weights. The comparison shows the effect directly: the grounded teacher
agrees with expert risk at $\kappa = 0.74$, whereas the same teacher run leakage-free
(stripped of that scaffolding) is correct on only 36\% of expert-set risk verdicts
(Section~\ref{sec:results-compression}). Finally, during
training, per-record loss masking is applied dynamically based on the
\texttt{tasks\_available} attribute. Expert-gold records supervise only hazard and risk
prediction, whereas expert-silver and street-view records provide supervision across all
three tasks.

\subsection{Student model and low-rank adaptation}
The student is Qwen3-VL-8B-Instruct \citep{qwen2025} with the vision encoder frozen. We
adapt the language backbone with LoRA \citep{hu2022} of rank $r = 16$, scaling
$\alpha = 32$, and dropout 0.05 on the attention projections ($q,k,v,o\_proj$), training
in bf16 with gradient checkpointing. Input images are processed at 1024-px native
resolution (the maximum dimension, so no information is discarded), a setting selected by
a zero-shot resolution probe. Freezing the vision encoder and adapting only low-rank
attention updates keeps the trainable footprint small enough for single-GPU fine-tuning
\citep{dettmers2023}.

\subsection{Training objective and imbalance handling}
\label{sec:method-loss}
Training minimises a weighted sum of per-task normalised cross-entropy losses. For a
record with supervised task set $T$, the loss is
\begin{equation}
\mathcal{L} = \sum_{t \in T} w_t \, \mathrm{CE}_t,
\qquad
\mathrm{CE}_t = \frac{1}{|S_t|}\sum_{i \in S_t} -\log p_\theta(y_i \mid y_{<i}, x),
\label{eq:loss}
\end{equation}
where $S_t$ is the set of target tokens for task $t$, $x$ is the (image, prompt) input,
and the task weights are $w_{\text{hazard}}=1.0$, $w_{\text{risk}}=1.0$, and
$w_{\text{rec}}=0.5$. Normalising within each task's supervised tokens keeps the short
risk target from being drowned by the longer hazard and recommendation text.

To handle the risk imbalance (Section~\ref{sec:data-splits}) we apply train-only logit
adjustment \citep{menon2021} on the three risk-token logits, adding a prior-dependent
offset at temperature $\tau=1$:
\begin{equation}
\tilde{z}_c = z_c + \tau \log \pi_c,
\qquad
\pi_c = \frac{n_c}{\sum_{c'} n_{c'}},
\label{eq:logitadj}
\end{equation}
with class counts $(n_{\text{Low}}, n_{\text{Med}}, n_{\text{High}}) =
(225, 6{,}340, 9{,}517)$. At $\tau=1$ this is Fisher-consistent for balanced error and is
equivalent to balanced softmax \citep{ren2020}. We use it instead of hand-set or
inverse-frequency class weights (the latter considered and rejected; \citealp{cui2019})
and do not stack a resampler. The adjustment is train-only: at inference we decode from
the raw logits, and any operating-point shift is handled post hoc
(Section~\ref{sec:method-inference}).

The proposed rebalancing strategy is intentionally restricted to the risk prediction
head, which consists of a single softmax over a small set of mutually exclusive classes.
This corresponds to the long-tailed multiclass classification setting for which
class-balanced loss, balanced meta-softmax, and logit adjustment were originally designed
\citep{cui2019,ren2020,menon2021}. In contrast, we do not rebalance the hazard prediction
head. Unlike the risk head, hazard prediction is formulated as a variable-length,
multi-label sequence generation task in which the autoregressive decoder produces
multiple multi-token hazard category spans rather than selecting a single class from a
fixed vocabulary. Consequently, training follows the standard visual instruction tuning
paradigm using token-level cross-entropy loss \citep{liu2023}, for which a single
class-specific reweighting factor is not naturally defined. Furthermore, our analysis in
Section~\ref{sec:data-analysis} shows that hazard categories with lower performance are
primarily constrained by their visual inferability rather than by their frequency in the
training data (Fig.~\ref{fig:baserate}b). Therefore, applying frequency-based rebalancing
to the hazard distribution would not be expected to improve recognition of these
categories. For this reason, the hazard prediction head is trained using the empirical
category distribution without additional reweighting.

\subsection{Optimisation and schedule}
We train for 2 epochs at learning rate $1\times10^{-4}$ with a cosine schedule and 3\%
warmup, an effective batch size of 16 (micro-batch 2 $\times$ gradient accumulation 8),
and early stopping on validation QWK (Table~\ref{tab:trainconfig}). Over the two epochs
the three task losses all decreased, risk-token accuracy rose, and validation QWK improved
to the selected checkpoint ($\approx 0.52$); the corresponding curves are reported as an
outcome in Section~\ref{sec:results-training}. Training used a single NVIDIA A100 40\,GB
GPU and about 6.3 hours of compute.

\begin{table}[t]
\centering
\caption{Training configuration.}
\label{tab:trainconfig}
\small
\begin{tabular}{@{}ll@{}}
\toprule
\textbf{Component} & \textbf{Setting} \\
\midrule
Base model            & Qwen3-VL-8B-Instruct (vision encoder frozen) \\
Adaptation            & LoRA, $r=16$, $\alpha=32$, dropout 0.05, on $q,k,v,o\_proj$ \\
Precision             & bf16 + gradient checkpointing \\
Image resolution      & 1024\,px (native max dimension) \\
Optimiser schedule    & LR $1\times10^{-4}$, cosine, 3\% warmup \\
Effective batch       & 16 (micro 2 $\times$ accumulation 8) \\
Epochs                & 2, early stop on validation QWK \\
Task weights          & hazard 1.0 / risk 1.0 / recommendation 0.5 \\
Imbalance handling    & train-only logit adjustment, $\tau=1$ \\
Decoding              & max\_new\_tokens 1024, stop sequences at end-of-audit \\
Compute               & single NVIDIA A100 40\,GB, $\sim$6.3\,h \\
\bottomrule
\end{tabular}
\end{table}

\subsection{Inference and operating-point selection}
\label{sec:method-inference}
At inference, the model generates structured JSON outputs with max\_new\_tokens $=1024$,
which comfortably exceeds the 99th percentile of target sequence lengths and prevents
audit truncation. Generation is terminated using stop sequences once the JSON output is
complete, and in practice the model consistently finishes well before the token limit.
Batched inference is served using vLLM \citep{kwon2023}. We additionally provide a post-hoc operating-point control. On the
validation set we cache the raw risk-token logits and sweep an additive per-class offset
$\delta_c$ to maximise validation QWK,
\begin{equation}
\hat{y} = \arg\max_c \; \big(z_c + \delta_c\big),
\qquad
\delta = \arg\max_{\delta} \; \mathrm{QWK}_{\text{val}}(\delta),
\label{eq:delta}
\end{equation}
then freeze $\delta$ and apply it once at test. As reported in
Section~\ref{sec:results-honest}, the validation-optimal $\delta$ trades a small amount of
ordinal QWK for substantially higher Low recall, so it functions as a deployment-time
choice rather than a fixed setting.

\subsection{Evaluation protocol and metrics}
\label{sec:method-eval}
The headline metric for overall risk is the quadratic weighted kappa (QWK)
\citep{torre2018}, a chance-corrected agreement coefficient for ordinal targets. For the
$C=3$ ordered risk levels (Low $<$ Medium $<$ High), let $O$ be the $C \times C$ confusion
matrix ($O_{ij}$ counts test records of true level $i$ predicted as level $j$), let $E$ be
the matrix expected when the predicted and true labels are independent
($E_{ij} = \frac{1}{N} O_{i\cdot}\, O_{\cdot j}$, where $O_{i\cdot}$ and $O_{\cdot j}$ are
the row and column totals and $N$ is the number of records), and let the ordinal penalties
be $w_{ij} = (i-j)^2 / (C-1)^2$. Then
\begin{equation}
\mathrm{QWK} = 1 - \frac{\sum_{i,j} w_{ij}\, O_{ij}}{\sum_{i,j} w_{ij}\, E_{ij}}.
\label{eq:qwk}
\end{equation}
The quadratic weights make the extreme Low$\leftrightarrow$High confusion four times as
costly as an adjacent Medium$\leftrightarrow$High confusion ($w=1$ versus $w=1/4$ at
$C=3$), so QWK rewards predictions that stay ordinally close: $\mathrm{QWK}=1$ is perfect
agreement, $0$ is chance-level, and negative values are worse than chance. We also report
exact-risk accuracy, per-class precision/recall/F1 and macro-F1, and hazard-category
precision/recall/F1 over all 12 categories (\texttt{skid\_resistance} and
\texttt{drainage} included, framed per Section~\ref{sec:data-taxonomy}).

Every point estimate carries a non-parametric bootstrap 95\% confidence interval
\citep{efron1979}. Let the test set be $\mathcal{D} = \{r_1, \dots, r_N\}$ and let
$m(\cdot)$ denote any of the metrics above, with point estimate $\hat{m} = m(\mathcal{D})$.
For each replicate $b = 1, \dots, B$ we draw a resample
$\mathcal{D}^{*}_{b} = \{r^{*}_1, \dots, r^{*}_N\}$ by sampling $N$ records from
$\mathcal{D}$ uniformly with replacement and recompute the statistic,
$\theta^{*}_{b} = m(\mathcal{D}^{*}_{b})$. The confidence interval is the percentile
interval over the $B$ replicates,
\begin{equation}
\mathrm{CI}_{95\%}(m) = \Big[\, Q_{2.5}\big(\{\theta^{*}_{b}\}_{b=1}^{B}\big),\;
Q_{97.5}\big(\{\theta^{*}_{b}\}_{b=1}^{B}\big) \,\Big],
\label{eq:bootstrap}
\end{equation}
where $Q_{p}(\cdot)$ is the $p$-th empirical percentile; we use $B = 2{,}000$ resamples
(raised to $5{,}000$ for the small gold-only and pooled expert subsets of
Section~\ref{sec:results-honest}, where the sample is smaller). Source-stratified
intervals follow the same procedure, resampling within each provenance tier so that the
gold, silver, and street-view subsets each carry their own uncertainty. The comparison
models are evaluated under identical single-image, leakage-free prompts: the zero-shot
Qwen3-VL-8B-Instruct base (the un-fine-tuned ablation), the 31B teacher run leakage-free,
and the proprietary frontier model Gemini-2.5-Flash \citep{comanici2025}. A blind human evaluation by a domain
expert scores a rubric (issue correctness, action appropriateness, LGED grounding, hazard
completeness, overall quality, and a binary risk-correct judgement) with model identity
hidden and order randomised; this design follows directly from the judge-bias and
circularity concerns of Section~\ref{sec:rw-eval}, which also dictate running the teacher
leakage-free. Finally, an internal-consistency check compares predicted risk with the risk
implied by the predicted High-severity hazard count (the 0/1/$\geq$2 rule).

%==============================================================
\section{Results}
\label{sec:results}

\subsection{Experimental setup}
\label{sec:results-setup}
We evaluate on the full location-disjoint test set ($n=3{,}447$). All comparison models
--- the zero-shot Qwen3-VL-8B-Instruct base, the 31B teacher run leakage-free, and
Gemini-2.5-Flash --- run under identical single-image, leakage-free prompts, with metrics
as defined in Section~\ref{sec:method-eval}. Generation was reliable: a risk token was
located in 100\% of the 3{,}447 generations, and parsed text-versus-logit agreement was
0.98.

\subsection{Training dynamics}
\label{sec:results-training}
\begin{figure}[t]
\centering
\includegraphics[width=\linewidth]{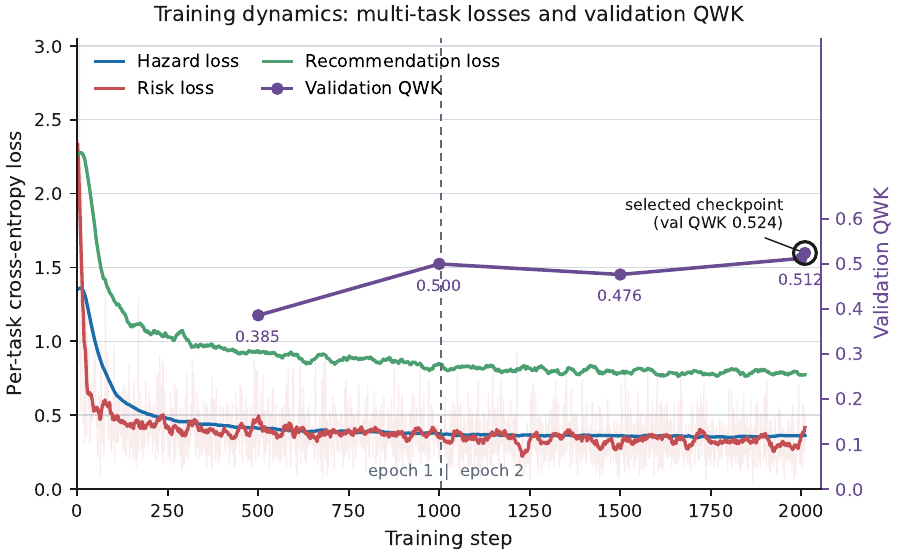}
\caption{Training dynamics over the full run: per-task losses (hazard, risk,
recommendation) and validation QWK against training steps, with the selected checkpoint
marked. All task losses decrease and validation QWK rises across the two epochs.}
\label{fig:training}
\end{figure}

Figure~\ref{fig:training} shows the training outcome. All three task losses fall over the
two epochs (hazard $1.34 \rightarrow 0.36$, risk $2.69 \rightarrow \approx 0.30$,
recommendation $2.23 \rightarrow 0.78$), confirming that the multi-task masking works and,
in particular, that the short risk target learns rather than being drowned by the longer
text. Risk-token accuracy rises from 0.56 to about 0.90, and validation QWK climbs from
0.385 to 0.524, which selects the early-stop checkpoint.

\subsection{Overall-risk prediction: fine-tuned versus zero-shot base}
\label{sec:results-headline}
\begin{figure}[t]
\centering
\includegraphics[width=0.85\linewidth]{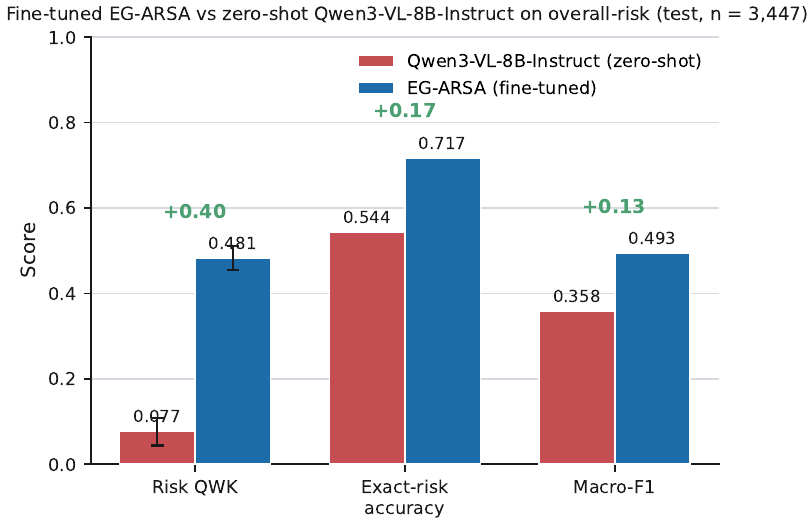}
\caption{Risk QWK and exact-risk accuracy for the zero-shot Qwen3-VL-8B-Instruct base
versus the fine-tuned student, visualising the $+0.40$ QWK gain from Expert-Grounded
fine-tuning.}
\label{fig:basevsft}
\end{figure}

\begin{figure}[t]
\centering
\includegraphics[width=0.9\linewidth]{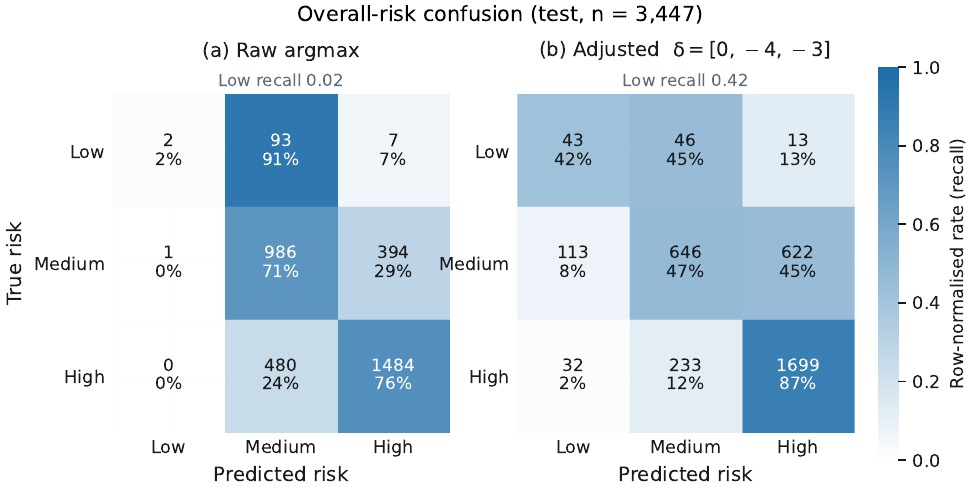}
\caption{Overall-risk confusion matrices for the fine-tuned student: raw operating point
(left) and $\delta$-adjusted (right). Errors are predominantly between adjacent risk
levels; the $\delta$-adjusted matrix recovers Low recall at a small ordinal cost.}
\label{fig:confusion}
\end{figure}

Fine-tuning produces a decisive gain over the zero-shot base (Table~\ref{tab:headline}).
At its default (raw) operating point, the fine-tuned student reaches risk QWK $0.4815$
($[0.4538, 0.5097]$), with linear $\kappa = 0.4562$, macro-F1 $0.4933$, and exact-risk
accuracy $0.7171$; all headline figures in this paper refer to this raw operating point,
and the post-hoc $\delta$-adjusted variant (Section~\ref{sec:results-honest}) is reported
only as an optional deployment alternative. The zero-shot base reaches QWK $0.0772$
($[0.0440, 0.1082]$) at accuracy $0.5440$ and never predicts Low (Low F1 $=0$).
Fine-tuning thus adds $+0.40$ QWK and $+0.17$ accuracy, and the bootstrap confidence
intervals do not overlap. The per-class breakdown (Table~\ref{tab:perclass}) shows strong
High and Medium performance and the known Low-recall limitation of the raw operating point,
which Section~\ref{sec:results-honest} revisits. Figure~\ref{fig:basevsft} visualises the
gain, and the confusion matrix in Fig.~\ref{fig:confusion} shows that residual errors are
overwhelmingly adjacent (Medium$\leftrightarrow$High), as the ordinal metric rewards.
Accuracy is consistent across sources (overall 0.717; gold 0.759; silver 0.702;
street-view 0.716), and the gold subset shows the largest fine-tuning lift (from 0.27
zero-shot to 0.76).

\begin{table}[t]
\centering
\caption{Overall-risk prediction on the test set ($n=3{,}447$): the fine-tuned student
versus the zero-shot base. Brackets give bootstrap 95\% confidence intervals.}
\label{tab:headline}
\small
\begin{tabular}{@{}lcccc@{}}
\toprule
\textbf{Model / setting} & \textbf{QWK} & \textbf{Linear $\kappa$} &
\textbf{Macro-F1} & \textbf{Accuracy} \\
\midrule
Zero-shot base (Qwen3-VL-8B) & 0.0772 [0.0440, 0.1082] & 0.0831 & 0.3585 & 0.5440 \\
Fine-tuned (raw)             & \textbf{0.4815 [0.4538, 0.5097]} & 0.4562 & 0.4933 & \textbf{0.7171} \\
\bottomrule
\end{tabular}
\end{table}

\begin{table}[t]
\centering
\caption{Per-class precision/recall/F1 for the fine-tuned student (raw operating point) on
the test set.}
\label{tab:perclass}
\small
\begin{tabular}{@{}lrrrr@{}}
\toprule
\textbf{Risk class} & \textbf{Precision} & \textbf{Recall} & \textbf{F1} &
\textbf{Support} \\
\midrule
Low    & 0.667 & 0.020 & 0.038 & 102 \\
Medium & 0.632 & 0.714 & 0.671 & 1{,}381 \\
High   & 0.787 & 0.756 & 0.771 & 1{,}964 \\
\bottomrule
\end{tabular}
\end{table}

\subsection{Performance against the teacher and a frontier model}
\label{sec:results-compression}
The compression headline is in Table~\ref{tab:multimodel}, on the 262 expert-grounded
entries under identical leakage-free inference. (This expert-grounded subset is a
different, smaller population than the full test set of
Section~\ref{sec:results-headline}, so the student's risk accuracy here, 0.737, is not the
same quantity as its 0.717 full-test accuracy; both are reported with their populations.)
The 8B student is risk-correct 73.7\% of the time (gold 0.753 / silver 0.712), against
Gemini-2.5-Flash at 59.2\% (0.544 / 0.663) and the 31B teacher run leakage-free at 35.9\%
(0.266 / 0.500). The student therefore beats a frontier proprietary model by $+0.14$, which is a
circularity-free comparison, since Gemini never produced any training label. It also beats
its own 31B teacher by a wide margin. Experiments show that the unaided teacher is the
weakest of the three. On silver hazard-category detection the picture is a
recall/precision trade-off: the student attains recall/precision 0.65/0.85, against Gemini
0.79/0.77 and the teacher 0.72/0.86. So, the frontier model enumerates more hazards while
the student is more precise.

\begin{table}[t]
\centering
\caption{Multi-model comparison on the 262 expert-grounded entries, under identical
single-image, leakage-free prompts. Risk accuracy is reported overall and split by source;
silver hazard-category recall/precision is also shown.}
\label{tab:multimodel}
\small
\begin{tabular}{@{}lcccc@{}}
\toprule
& \multicolumn{3}{c}{\textbf{Risk accuracy}} & \textbf{Silver hazard R/P} \\
\cmidrule(lr){2-4}
\textbf{Model} & All & Gold & Silver & (recall / precision) \\
\midrule
EG-ARSA (8B student)      & \textbf{0.737} & 0.753 & 0.712 & 0.65 / 0.85 \\
Gemini-2.5-Flash          & 0.592 & 0.544 & 0.663 & 0.79 / 0.77 \\
Teacher (31B, leakage-free) & 0.359 & 0.266 & 0.500 & 0.72 / 0.86 \\
\bottomrule
\end{tabular}
\end{table}

\subsection{Blind human evaluation}
\label{sec:results-human}
A domain expert scored model outputs blind, with identity hidden and order randomised
(Table~\ref{tab:human}). On the 80-item blind comparison the student leads on overall
quality (3.94/5) and is risk-correct 81\% of the time ($[0.73, 0.90]$), against Gemini at
3.65 / 0.58 and the teacher at 3.46 / 0.42; the student's sub-dimension means are issue
correctness 3.99, action appropriateness 4.03, LGED grounding 4.95, and hazard
completeness 3.41. On the full expert-grounded set ($n=262$, all gold and silver), the
student's rubric means are issue 4.02, action 4.06, LGED grounding 4.92, hazard
completeness 3.48, and overall 3.97, with risk-correct 0.847 (222/262); 194 of 262 outputs
(74\%) score 4--5 overall.

The strongest evidence is that two independent measures agree: blind-human risk-correct and
fully automated risk accuracy reproduce the same ranking (student 0.81 / 0.74, Gemini
0.58 / 0.59, teacher 0.42 / 0.36). Because the human panel cannot be influenced by any
label-generation circularity and the automated metric is computed independently, their
agreement indicates the ranking is not a metric artifact.

\begin{table}[t]
\centering
\caption{Blind human evaluation. Top: the 80-item blind three-model comparison (overall
quality on a 1--5 scale and binary risk-correct). Middle: the student's full
expert-grounded evaluation ($n=262$). Bottom: the cross-validation contrasting
blind-human risk-correct with independent automated risk accuracy.}
\label{tab:human}
\small
\begin{tabular}{@{}lcc@{}}
\toprule
\multicolumn{3}{@{}l}{\emph{Blind three-model comparison ($n=80$)}} \\
\textbf{Model} & \textbf{Overall (1--5)} & \textbf{Risk-correct} \\
\midrule
EG-ARSA (8B student) & \textbf{3.94} & \textbf{0.81 [0.73, 0.90]} \\
Gemini-2.5-Flash     & 3.65 & 0.58 \\
Teacher (31B, leakage-free) & 3.46 & 0.42 \\
\midrule
\multicolumn{3}{@{}l}{\emph{EG-ARSA full expert-grounded evaluation ($n=262$)}} \\
Issue correctness & \multicolumn{2}{c}{4.02} \\
Action appropriateness & \multicolumn{2}{c}{4.06} \\
LGED grounding & \multicolumn{2}{c}{4.92} \\
Hazard completeness & \multicolumn{2}{c}{3.48} \\
Overall quality & \multicolumn{2}{c}{3.97} \\
Risk-correct & \multicolumn{2}{c}{0.847 (222/262)} \\
\midrule
\multicolumn{3}{@{}l}{\emph{Cross-validation: blind-human vs. automated risk}} \\
\textbf{Model} & \textbf{Human risk-correct} & \textbf{Automated risk acc.} \\
\midrule
EG-ARSA (8B student) & 0.81 & 0.74 \\
Gemini-2.5-Flash     & 0.58 & 0.59 \\
Teacher (31B, leakage-free) & 0.42 & 0.36 \\
\bottomrule
\end{tabular}
\end{table}

\subsection{Internal consistency and reliability}
\label{sec:results-consistency}
On the street-view test records, the model's risk accuracy (0.716,
Section~\ref{sec:results-headline}) is on par with the 72.2\% agreement that the
severity-count rule itself attains against the expert labels
(Section~\ref{sec:data-splits}): the risk head reaches the accuracy ceiling a counting
heuristic achieves on this data. Yet the model's predicted risk coincides with that same
0/1/$\geq$2 rule applied to its \emph{own} predicted High-severity hazards only 48.4\% of
the time. This self-consistency figure measures how far the risk head departs from a
mechanical count of its own hazard list, not its agreement with ground truth. Taken
together, the risk head matches counting-level accuracy against the experts while not
simply counting; it integrates scene context, as intended. Combined with the 100\%
risk-token location rate and 0.98 parsed-versus-logit agreement
(Section~\ref{sec:results-setup}), the outputs are reliable to parse and score.

\subsection{External validity}
\label{sec:results-honest}
On the gold-only and pooled expert subsets, QWK against the teacher reference
($\kappa=0.74$) is near zero, but this is a property of the subset, not the model. The
gold subset is 82\% High (130/158, with only 2 Low cases), so QWK is degenerate on a
near-single-class sample. The informative metric there is accuracy, and gold accuracy is
the highest of any source (0.759, with within-one-ordinal accuracy 0.987). Accordingly,
the claim that the student beats the teacher rests on the leakage-free multi-model
comparison and the blind human evaluation
(Sections~\ref{sec:results-compression}--\ref{sec:results-human}), not on gold QWK.

Secondly, the raw operating point used for all headline numbers maximises ordinal QWK but
is deliberately conservative on the minority Low class (recall 0.02). For deployments that
need Low-risk sensitivity, the post-hoc per-class offset $\delta$ provides a lever: applied
once at test it recovers Low recall to 0.42 (F1 0.30) at a small ordinal-QWK cost
(Fig.~\ref{fig:confusion}). It is therefore a deployment-time operating-point choice, not
part of the headline result.

%==============================================================
\section{Discussion}
\label{sec:discussion}

\subsection{Why Expert-Grounded Distillation works}
The compression result is not about scale; it is about grounding. The student internalises
expert-grounded audit knowledge and deploys it from a single image, whereas the teacher
unaided collapses (teacher--expert $\kappa$ 0.74 grounded $\rightarrow$ 0.36 leakage-free)
and is the weakest of the three models even though it generated the bulk of the training
labels. This dissociation, the label generator auditing worst when it must work alone,
is the core interpretation. What the task needs is the expert grounding, and EGD is the
mechanism that moves that grounding into a small, deployable model. The outcome is precisely
the small-student-beats-large-model result anticipated by \citet{hsieh2023} and
\citet{zhong2025}, and the kind of improvement beyond the generator that \citet{amin2025}
show is reachable from curated, human-reviewed ``weak'' data. It also closes the gap
identified in Section~\ref{sec:related}: prior VLM auditing was zero-shot and proprietary,
and grounded fine-tuning is what moves the paradigm forward. EGD builds on established
mechanisms such as context distillation \citep{askell2021,snell2022} and weak-to-strong
generalization \citep{burns2023}. It then turns them into a working method for a real
engineering domain. It is, to our knowledge, the first pipeline to make expert grounding a
quantified, gated step and to distill institutional field-audit expertise into a compact
open auditor, delivered together with the first open dataset and model for LMIC
road-safety auditing.

\subsection{Deployment, interpretability, and cost}
EG-ARSA outputs an interpretable language audit with concrete recommendations, not just a
score, which makes it actionable for non-expert users and deployable through a web
application. Because $\delta$ is a single frozen post-hoc offset, a deployment can choose
its operating point (Low recall for screening or ordinal QWK for ranking) without
retraining. The cost argument is central to the LMIC motivation: the student was fine-tuned
on a single A100 in about six hours and runs on modest hardware, making per-kilometre audit
coverage orders of magnitude cheaper than a formal field RSA. This directly addresses the
coverage-versus-cost gap quantified by \citet{li2024} in exactly the setting where formal
audits are unaffordable.

\subsection{Generalizability}
The EGD pipeline is jurisdiction-agnostic: institutional-audit grounding distilled into a
compact open student transfers to any setting with even a small expert-audited corpus. The
near-national street-view footprint (63 districts, 8 divisions) already shows the inference
side spanning diverse rural and suburban conditions, and the approach extends naturally to
other LMICs with iRAP or RSA programmes.

\subsection{Limitations}
\label{sec:lim}
EG-ARSA audits a single street-view image, so full road geometry and the non-visual
extremes are recoverable only in obvious cases. The street-view labels are
teacher-generated and, although produced by an expert-calibrated prompt and partly
human-reviewed, inherit some of the teacher's ceiling, which is why the expert-grounded
gold and silver tiers anchor evaluation. The expert ground truth is concentrated in the 18
LGED-audited districts, while the nationwide spread is teacher-labelled street-view;
broadening the expert-grounded footprint is future work. The system targets the
rural/suburban LGED road class, with national highways (RHD) and city streets (City
Corporations) under other jurisdictions.

\subsection{Future work}
A complementary overhead or satellite stream could supply the geometry and non-visual
attributes that a single street-view image cannot, and learned fusion of street and
overhead views is a promising extension. Collaboration with RHD and the City Corporations
would extend the same EGD recipe to highways and city streets, building towards a
comprehensive national auditor across all road classes. We also see value in broadening the
expert-grounded corpus to more districts, in richer recommendation generation, and in
abstention on borderline single-image cases.

%==============================================================
\section{Conclusion}
\label{sec:conclusion}
We addressed scalable road safety auditing for low-resource settings, where reliable crash
records are often unavailable and expert-led infrastructure audits cannot be performed at
national scale. We introduced \emph{Expert-Grounded Distillation} (EGD), a framework that
grounds teacher VLM supervision in authoritative institutional field audits, validates the
generated supervision through human review, and distills this expertise into a compact open
model (\textbf{EG-ARSA}) using a single leakage-free prompt. We also release
\textbf{BD-ARSA}, the first open, expert-grounded Bangladeshi road safety visual-audit
dataset. Grounded fine-tuning improves the student model's ordinal risk agreement by
$+0.40$ quadratic weighted kappa over its zero-shot baseline, while blind expert evaluation
shows that the compact 8B model outperforms both its 31B teacher and a frontier proprietary
model, with automated metrics reproducing the same ranking. These results demonstrate that
expert-grounded supervision can outperform raw model scale, enabling accurate, affordable,
and deployable road safety auditing where conventional Road Safety Audits are impractical.
Future work will extend EGD to additional road classes through the Roads and Highways
Department (RHD) and City Corporations, and incorporate complementary sensing modalities
toward a nationwide road safety auditing framework.

%==============================================================
\section*{CRediT authorship contribution statement}
\textbf{Md Thamed Bin Zaman Chowdhury:} Conceptualization, Methodology, Software,
Validation, Formal analysis, Investigation, Data curation, Writing --- original draft,
Visualization. \textbf{Moazzem Hossain:} Methodology, Validation, Writing --- review \&
editing, Supervision, Project administration.

\section*{Data and code availability}
The BD-ARSA dataset, the fine-tuned model (LoRA adapter), and the training and evaluation
code with reproduction steps are released openly:
\begin{itemize}
  \item \textbf{Dataset} --- audit annotations and metadata (CC~BY~4.0); the Google
  Street View imagery is not redistributed and is reconstructed locally via the included
  fetch script:\
  \url{https://huggingface.co/datasets/Thamed-Chowdhury/bd-arsa-road-safety-visual-audit}
  \item \textbf{Model} --- Apache-2.0 LoRA adapter for Qwen3-VL-8B-Instruct; use is
  additionally subject to the Gemma Terms of Use, as the distillation supervision was
  Gemma-generated:\
  \url{https://huggingface.co/Thamed-Chowdhury/eg-arsa-qwen3vl-8b-lora}
  \item \textbf{Code} --- Apache-2.0:\
  \url{https://github.com/Thamed-Chowdhury/EG-ARSA}
\end{itemize}

All test-set results in this
paper are computed on the frozen, location-disjoint test split ($n=3{,}447$); the
multi-model and human evaluations use its full expert-grounded subset ($n=262$: 158
expert-gold $+$ 104 expert-silver records), and the silver hazard-category overlap
(Section~\ref{sec:results-compression}) is computed over the 557 expert findings in the
104 silver test records.

\paragraph{Data provenance and licensing} The released annotations, audit labels, scene descriptions, recommendations,
capture coordinates, and panorama identifiers are licensed CC~BY~4.0. The street-view
imagery is \copyright~Google: it was accessed through the Google Maps Platform under its
Terms of Service and is not redistributed, but is reconstructed locally from the released
panorama identifiers and coordinates through the official Street View Static API using the
included fetch script. The expert-gold and expert-silver images derive from the
ARI--BUET/LGED field-audit reports and are shared with LGED's written permission. The
street-view teacher supervision was generated with Google's Gemma model; consequently the
released model is additionally subject to the Gemma Terms of Use and Prohibited Use
Policy. The original ARI--LGED PDF reports can be shared for review purposes.

\section*{Acknowledgements}
The expert ground truth derives from on-site Road Safety Audits conducted by faculty of the
Accident Research Institute (ARI), Bangladesh University of Engineering and Technology
(BUET), commissioned by the Local Government Engineering Department (LGED) under the World
Bank--financed Second Rural Transport Improvement Project (RTIP-II, Additional Financing;
P166295). We express our sincere gratitude to LGED for giving us permission to use and share
this data for this research.

%==============================================================
\bibliographystyle{elsarticle-harv}
\bibliography{references}

%==============================================================
\appendix

\section{NLP pipeline for taxonomy derivation and corpus validation}
\label{app:nlp}
This appendix documents the natural-language-processing pipeline that turned the raw expert
audit text into the 12-category BD-ARSA taxonomy, together with the analyses used to
validate it. All analyses operate on the expert finding corpus extracted from the ARI--LGED
audit reports.

\subsection{Deriving the 12-category schema with dual zero-shot classifiers}
The expert reports record hazards as free-text findings. We collected 689 unique finding
texts (208 distinct short labels) and mapped each to one of the 12 LGED categories using two
independent zero-shot classifiers --- Gemini-2.5-Flash and BART-large-MNLI (zero-shot
natural-language inference). The two classifiers agreed only moderately
(Table~\ref{tab:interclass}); we did not treat this as a reliability estimate for the
taxonomy but used the 82 label-level disagreements to route ambiguous findings to human
adjudication. The final mapping carries a per-item confidence (high 84 / medium 42 / low 82
short labels), and 134 hazard-image labels whose category remained ambiguous were
quarantined from the gold tier.

\begin{table}[t]
\centering
\caption{Inter-classifier agreement between the two zero-shot label generators
(Gemini-2.5-Flash vs. BART-large-MNLI).}
\label{tab:interclass}
\small
\begin{tabular}{@{}lc@{}}
\toprule
\textbf{Comparison} & \textbf{Value} \\
\midrule
Short labels --- exact match (top-1 vs. top-1) & 40.4\% \\
Short labels --- top-2 inclusion & 60.6\% \\
Short labels --- Cohen's kappa & 0.331 \\
Finding texts --- mean Jaccard (multi-label) & 0.373 \\
Finding texts --- perfect agreement & 12.3\% \\
Finding texts --- low agreement (Jaccard $<$ 0.5) & 61.5\% \\
Label disagreements routed to human review & 82 / 208 \\
\bottomrule
\end{tabular}
\end{table}

\subsection{Corpus lexical analysis}
\begin{figure}[t]
\centering
\includegraphics[width=0.8\linewidth]{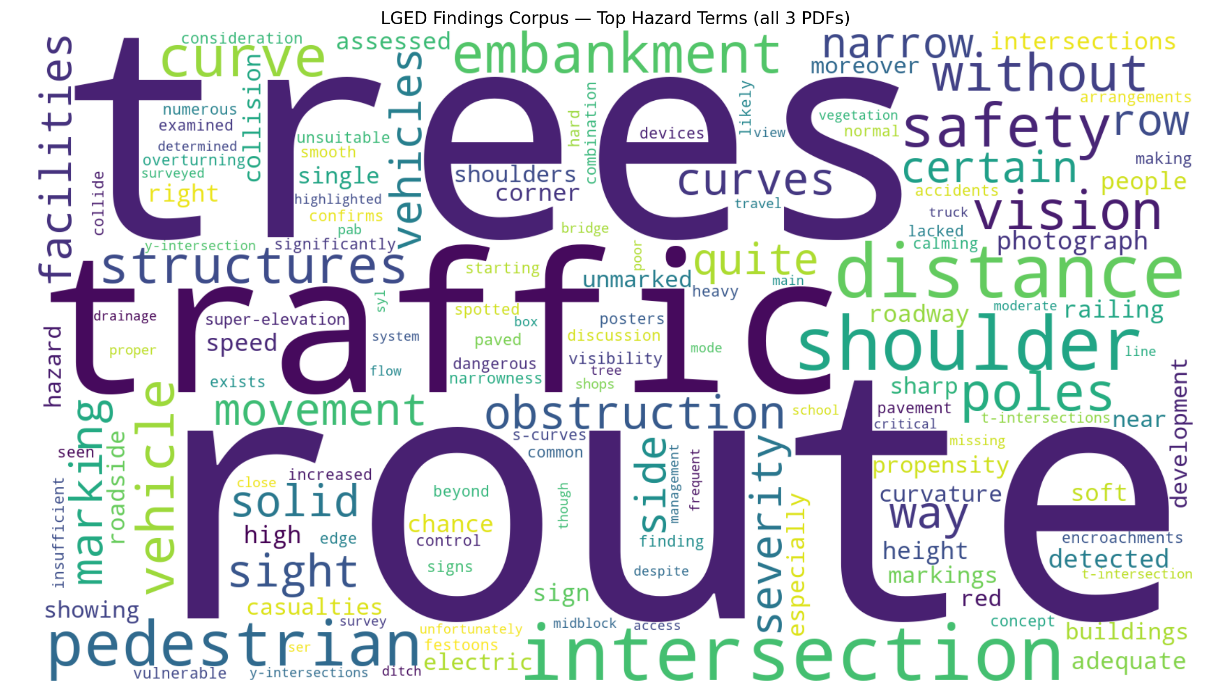}\\[4pt]
\includegraphics[width=0.8\linewidth]{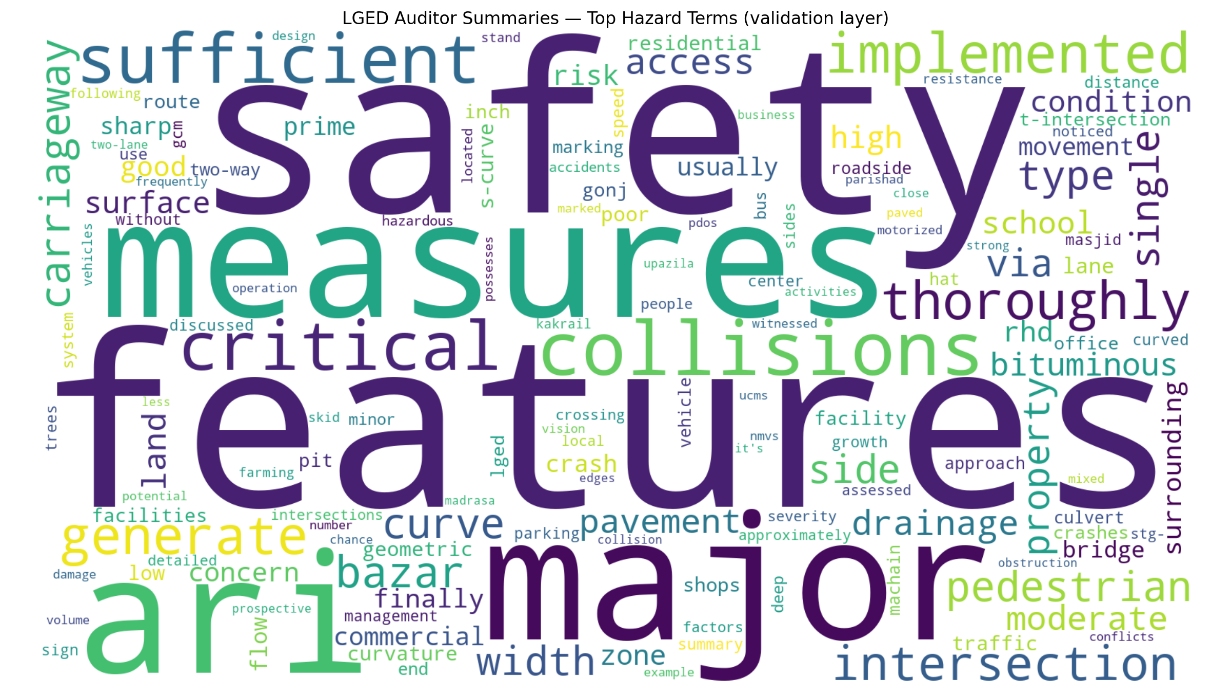}
\caption{Word clouds of the expert corpus: (top) the detailed per-location findings layer,
and (bottom) the independently-written auditor summary layer.}
\label{fig:wordclouds}
\end{figure}

Beyond classification, we profiled the corpus lexically to characterise its vocabulary and
to cross-check the taxonomy against an independent text source. The dominant unigrams of the
detailed-findings layer --- route (685), trees (459), traffic (414), intersection (382),
shoulder (360), distance (357), pedestrian (334), safety (298), curve (274), embankment
(262) --- track the hazard categories directly; Figure~\ref{fig:wordclouds} shows the
corresponding word clouds. As an independent witness, we compared the detailed per-location
findings against the separately-written auditor summaries. Although their surface n-grams
diverge (the two layers are written in different registers), both independently emphasise
the same hazard categories (Table~\ref{tab:keyword}), corroborating the category prominence
that the taxonomy encodes.

\begin{table}[t]
\centering
\caption{Category keyword signal: keyword hits per hazard category in the detailed-findings
layer and in the auditor-summary layer. Both independent layers cover all 12 categories.}
\label{tab:keyword}
\small
\begin{tabular}{@{}lrr@{}}
\toprule
\textbf{Category} & \textbf{Findings-layer hits} & \textbf{Summary-layer hits} \\
\midrule
\texttt{vision\_obstruction}   & 1822 & 72 \\
\texttt{roadside\_severity}    & 1300 & 141 \\
\texttt{intersection}          & 1212 & 176 \\
\texttt{shoulder\_condition}   & 1037 & 113 \\
\texttt{embankment\_safety}    & 915  & 154 \\
\texttt{speed\_management}     & 890  & 61 \\
\texttt{traffic\_signs}        & 851  & 41 \\
\texttt{pedestrian\_facilities} & 674 & 144 \\
\texttt{road\_markings}        & 422  & 82 \\
\texttt{drainage}              & 198  & 96 \\
\texttt{skid\_resistance}      & 135  & 89 \\
\texttt{bus\_stoppage}         & 123  & 56 \\
\bottomrule
\end{tabular}
\end{table}

\section{Prompts}
\label{app:prompts}
The prompts below are reproduced verbatim (with box-drawing rules rendered as plain ASCII
separators). The teacher prompt (\ref{app:prompt-teacher}) generated the street-view tier;
the student prompts (\ref{app:prompt-gold}, \ref{app:prompt-full}) are the leakage-free
instructions used to fine-tune and serve EG-ARSA. The deliberate asymmetry between them ---
the teacher's elaborate, LGED-grounded, multi-image prompt versus the student's single
standardised, single-image instruction --- is the Expert-Grounded Distillation mechanism
described in Section~\ref{sec:method-task}. Expert-gold labels were extracted from the expert
records and used no generation prompt.

\subsection{Teacher street-view generation prompt (gemma-4-31b-it)}
\label{app:prompt-teacher}
Used to generate the street-view tier from multiple ground-level Street View screenshots plus
one satellite tile per location; outputs were validated against the 12-category taxonomy,
risk-boundary post-processed, and human-reviewed on a 384-record sample.

\begin{lstlisting}
You are an expert road safety auditor following the Bangladesh Local Government
Engineering Department (LGED) road safety audit methodology. Analyse the provided
road images and conduct a comprehensive road safety audit.

The images you receive include:
- Google Maps Street View screenshots: ground-level views at various headings
- A satellite/aerial screenshot: overhead view of the road layout

==============================================================
LGED ISSUE CATEGORIES - check ALL of the following:
==============================================================

1. Speed Management
   Look for: speed limit signs, speed humps, rumble strips, traffic calming.
   Flag: complete absence of any speed-reduction infrastructure.

2. Road Markings
   Look for: centreline, edge/fog line, zebra crossing, turning arrows.
   Flag: absent, faded, or non-existent markings on any road with mixed
   traffic, pedestrian activity, intersections, curves, or settled (residential
   /commercial/educational) frontage.
   ONLY skip flagging on isolated agricultural roads with no nearby settlement,
   intersections, or pedestrian activity. When in doubt, flag it.

3. Vision / Sight Obstruction
   Look for: walls, buildings, dense vegetation, parked vehicles blocking
   sight lines at curves or intersections.
   Flag: any object that reduces stopping or intersection sight distance.

4. Drainage System
   Look for: roadside drains, culverts, standing water, waterlogged shoulders,
   road-edge erosion.
   Flag: blocked/absent drains, visible water ponding, erosion at carriageway edge.

5. Traffic Signs
   Look for: curve-ahead warnings, intersection warnings, school-zone signs,
   speed limit plates, directional/regulatory signs.
   Flag: missing or inadequate signage.

6. Skid Resistance
   Look for: polished/worn surface, bleeding asphalt, loose aggregate.
   Flag: surfaces likely to cause skidding in wet conditions.

7. Roadside Severity (Clear Zone)
   Look for: utility poles, trees, walls, or other rigid obstacles within
   approximately 1 metre (3 feet) of the carriageway edge.
   Flag as HIGH SEVERITY: any rigid obstacle within 3 feet of road edge.

8. Embankment / Drop-off Safety
   Look for: road embankments or cut slopes, guard rails, safety barriers.
   Flag as HIGH SEVERITY: embankment height > 6 feet (2 m) without any
   safety railing - single-vehicle overturn risk.

9. Shoulder Condition
   Look for: paved shoulder, gravel/earth (soft) shoulder, effective width.
   IMPORTANT: Examine the satellite image carefully - it shows shoulder
   width from above far better than street-level views. Also scan every
   Street View image edge for the road/ground transition.
   Flag: complete absence of both paved and soft shoulders; effective
   width < 3 feet. Many Bangladesh rural roads have NO usable shoulder -
   do not omit this category just because it is hard to see from Street View.
   If you cannot confirm a usable shoulder >= 3 ft wide in the satellite view,
   flag it.

10. Pedestrian Facilities
    Look for: dedicated footpath, pedestrian crossings, guardrails separating
    pedestrians from traffic; observe apparent pedestrian activity level.
    Flag: high pedestrian flow with no dedicated facilities.

11. Bus Stoppage Standard
    Bangladesh context: rural bus/tempo/CNG stops are typically informal -
    no bay, no shelter, no marking. Buses stop wherever passengers congregate.
    Look for: VISIBLE evidence of regular stopping activity - e.g. people
    waiting/gathered at the roadside, an informal shelter/bench, a visible
    bus/tempo/CNG in the image, a busy market or school/college gate where
    public transport must stop, or worn carriageway edges from repeated stops.
    Flag ONLY when at least one of those indicators is visible AND no proper
    bay or marked stopping area exists. Do NOT flag every rural road; require
    concrete visual evidence of bus stopping activity at THIS location.

12. Intersection Quality
    Look for: intersection type (4-way, T-junction, staggered), approach
    visibility, channelisation, traffic-control devices.
    Flag ONLY when the intersection ITSELF has a primary structural/geometric
    defect visible in the images: (a) T-junction without flaring or turning
    provision, (b) blind approach where sight distance <= 30 m due to
    geometry (not vegetation - capture that under Vision Obstruction),
    (c) multiple uncontrolled entries at a busy junction with no channelling.
    IMPORTANT: If the only issue at a junction is poor sight distance caused
    by vegetation or structures, list it under Vision/Sight Obstruction ONLY
    - do NOT also create a separate Intersection Quality entry. Do not flag
    every minor junction; only those with a clearly observable geometric hazard.

==============================================================
ANALYSIS INSTRUCTIONS:
==============================================================

- Examine ALL provided images before forming conclusions.
- Report BOTH the presence of hazards AND the absence of required safety
  features (e.g. "No speed-management infrastructure observed").
- If Street View shows no imagery (grey/unavailable screen), state this clearly
  and rely primarily on the satellite image.
- Be specific: "utility pole ~1-2 feet from carriageway edge" is better than
  "pole near road".
- Consider the Bangladesh road context: mixed traffic (rickshaws, CNGs,
  buses, motorcycles, pedestrians sharing the carriageway).

==============================================================
OUTPUT FORMAT - return ONLY a raw JSON object, no markdown fences:
==============================================================

{
  "location_description": "<2-3 sentences: road type, setting, apparent traffic and pedestrian volume>",
  "road_type": "<single carriageway / dual carriageway / other>",
  "land_use": "<residential / commercial / educational / agricultural / mixed>",
  "critical_safety_issues": [
    "<Category Name - specific observation from the images>",
    "<Category Name - specific observation from the images>"
  ],
  "hazard_illustrations": [
    {
      "hazard_name": "<hazard category name>",
      "observation": "<what is specifically visible in the images>",
      "severity": "<High / Medium / Low>"
    }
  ],
  "overall_risk_level": "<High / Medium / Low>",
  "street_view_available": <true / false>,
  "additional_recommendations": "<any supplementary observations>"
}
\end{lstlisting}

\subsection{Student single-hazard prompt (expert-gold crops)}
\label{app:prompt-gold}
Applied to single-hazard crops; identifies the hazard and the location's overall risk from
one image, with no report context.

\begin{lstlisting}
You are a road-safety auditor inspecting rural roads in Bangladesh. This image is a
close-up that highlights a single road-safety hazard at a location. Identify that hazard
and the location's overall risk level, based only on what is visible. If a clear
remedial measure applies, also give a brief recommendation.

Respond with ONLY a JSON object in this schema (omit "additional_recommendations" if no
clear fix applies):
{
  "hazards": [
    {
      "hazard_name": "<short hazard name>",
      "hazard_category": "<one of: road_markings, shoulder_condition, roadside_severity, pedestrian_facilities, vision_obstruction, traffic_signs, speed_management, intersection, embankment_safety, bus_stoppage, drainage, skid_resistance>"
    }
  ],
  "overall_risk_level": "<Low | Medium | High>",
  "additional_recommendations": "<brief recommended improvement(s)>"
}
\end{lstlisting}

\subsection{Student full-audit, leakage-free prompt (expert-silver and street-view)}
\label{app:prompt-full}
The single standardised instruction used for full structured audits at both training and
inference; it carries no grounding context, no per-category corrective rules, and no report
references.

\begin{lstlisting}
You are a road-safety auditor inspecting rural and semi-urban roads in Bangladesh.
Carefully examine the road image and produce a structured road-safety audit based only
on what is visible in the image. Identify each safety hazard, describe what you see,
rate its severity, and give the overall risk for this location.

Respond with ONLY a JSON object in exactly this schema:
{
  "location_description": "<1-2 sentences describing the road scene>",
  "road_type": "<e.g. single carriageway>",
  "land_use": "<agricultural | residential | commercial | mixed | institutional>",
  "hazards": [
    {
      "hazard_name": "<short hazard name>",
      "hazard_category": "<one of: road_markings, shoulder_condition, roadside_severity, pedestrian_facilities, vision_obstruction, traffic_signs, speed_management, intersection, embankment_safety, bus_stoppage, drainage, skid_resistance>",
      "observation": "<what is specifically visible in the image>",
      "severity": "<Low | Medium | High>"
    }
  ],
  "critical_safety_issues": ["<short phrase>", "..."],
  "overall_risk_level": "<Low | Medium | High>",
  "additional_recommendations": "<concrete fixes for the visible deficiencies>"
}

Overall risk guide: High = several and/or serious hazards with a real chance of a crash
or injury; Medium = some hazards needing attention; Low = generally safe, only minor
issues.
\end{lstlisting}

\end{document}